\documentclass[a4paper,fleqn]{cas-sc}

\usepackage[numbers,sort&compress]{natbib}

\usepackage{scrextend}
\usepackage{graphicx}
\usepackage{xcolor}
\usepackage{amssymb}
\usepackage{multirow}
\usepackage{rotating}
\usepackage{url}
\usepackage[utf8]{inputenc}
\inputencoding{utf8}
\usepackage{soul}
\usepackage[ruled,vlined]{algorithm2e}
\usepackage{fancyhdr}

\def\tsc#1{\csdef{#1}{\textsc{\lowercase{#1}}\xspace}}
\tsc{WGM}
\tsc{QE}
\tsc{EP}
\tsc{PMS}
\tsc{BEC}
\tsc{DE}

\begin{document}

\fancypagestyle{firstpage}
{
    \fancyhf{}

    \fancyhead[L]{\small
    Cite as: Dazeley, R., Vamplew, P., Foale, C., Young, C., Aryal, S., \& Cruz, F. (2021). Levels of explainable artificial intelligence for human-aligned conversational explanations. Artificial Intelligence, 299, 103525. DOI: 10.1016/j.artint.2021.103525}

    \fancyfoot[L]{\small \textcopyright 2021. Licensed under the Creative Commons CC-BY-NC-ND 4.0 http://creativecommons.org/licenses/by-nc-nd/4.0/ 
    }

}

\let\WriteBookmarks\relax
\def\floatpagepagefraction{1}
\def\textpagefraction{.001}
\shorttitle{Levels of XAI}
\shortauthors{R Dazeley et~al.}

\title [mode = title]{Levels of Explainable Artificial Intelligence for Human-Aligned Conversational Explanations}                      



\newcommand{\Authornote}[2]{\small\textcolor{red}{\sf$[${#1: #2}$]$}}
\newcommand{\sanote}[2]{{\Authornote{Sunil}{#1}}} 
\newcommand{\rpdnote}[2]{{\Authornote{Richard}{#1}}} 

\newcommand{\edit}[1]{\textcolor{red}{#1}}
\newcommand{\cut}[1]{\textcolor{blue}{{#1}}}

\author[1]{Richard Dazeley}[
                        orcid=0000-0002-6199-9685]
\cormark[1]
\ead{richard.dazeley@deakin.edu.au}

\address[1]{School of Information Technology, Deakin University, Locked Bag 20000, Geelong, Victoria 3220, Australia}

\author[2]{Peter Vamplew}[orcid=0000-0002-8687-4424]
\ead{p.vamplew@federation.edu.au}
\author[2]{Cameron Foale}[orcid=0000-0003-2537-0326]
\ead{c.foale@federation.edu.au}
\author[2]{Charlotte Young}
\ead{cm.young@federation.edu.au}
\author[1]{Sunil Aryal}[orcid=0000-0002-6639-6824]
\ead{sunil.aryal@deakin.edu.au}
\author[1]{Francisco Cruz}[orcid=0000-0002-1131-3382]
\ead{francisco.cruz@deakin.edu.au}


\address[2]{School of Engineering, Information Technology and Physical Sciences, Federation University, Ballarat, Victoria 3353, Australia}



\begin{abstract}
Over the last few years there has been rapid research growth into eXplainable Artificial Intelligence (XAI) and the closely aligned Interpretable Machine Learning (IML). Drivers for this growth include recent legislative changes and increased investments by industry and governments, along with increased concern from the general public. People are affected by autonomous decisions every day and the public need to understand the decision-making process to accept the outcomes. However, the vast majority of the applications of XAI/IML are focused on providing low-level `narrow' explanations of how an individual decision was reached based on a particular datum. While important, these explanations rarely provide insights into an agent's: beliefs and motivations; hypotheses of other (human, animal or AI) agents' intentions; interpretation of external cultural expectations; or, processes used to generate its own explanation. Yet all of these factors, we propose, are essential to providing the explanatory depth that people require to accept and trust the AI's decision-making. This paper aims to define levels of explanation and describe how they can be integrated to create a human-aligned conversational explanation system. In so doing, this paper will survey current approaches and discuss the integration of different technologies to achieve these levels with \textit{Broad eXplainable Artificial Intelligence (Broad-XAI)}, and thereby move towards high-level `strong' explanations.
\end{abstract}




\maketitle


\thispagestyle{firstpage}

\section{Introduction}

Knowledge-Based Systems (KBS) researchers and designers have long understood that the ability of a system to explain its decisions is critical to human acceptance, with approaches to providing explanations having been discussed as early as \citeauthor{shortliffe1975model} \cite{shortliffe1975model} and later in a range of projects such as \cite{davis1977production, swartout1983xplain, chandrasekaran1988explanation}. This early body of work has been further applied in domains such as Bayesian Networks \citep{lacave2002review}, early Neural Network systems \citep{andrews1995survey} and in Recommender systems \citep{cramer2008effects, assad2007personisad}. However, Machine Learning systems developed this century, such as Deep Learning, have become increasingly obfuscated and non-transparent to users. 

Recently there has been a growth in interest in eXplainable Artificial Intelligence (XAI) and Interpretable Machine Learning (IML)\footnote{The literature has also occasionally used terms such as Transparent AI/ML and Explainable Machine Learning interchangeably with XAI and IML \cite{goyal2016towards, wachter2017transparent, chao2010transparent}.} \citep{abdul2018trends, adadi2018peeking}. One major driver for current research has been the XAI project launched by the Defense Advanced Research Projects Agency (DARPA), with twelve research programs receiving USD 75 million in funding \citep{kuang2017nytimes} to create a suite of explainable machine learning techniques \citep{gunning2018DARPA}. The DARPA Project is wide ranging with aims to develop both models and interfaces for explainability. The need for XAI has been further driven by governments beginning to legislate requirements for autonomous systems to provide explanations of their decisions. For instance, the European Union's new General Data Protection Regulation \citep{goodman2016european} requires autonomous systems to be able to provide explanations of any decisions that are based on an individual's data. As autonomous systems increase in their level of societal integration these legislative requirements for explanation are likely to increase. There has also been significant interest in XAI from futurists and innovation based companies such as AGI Innovations \citep{voss2018AGIInnovations} and bons.ai \citep{hammond2018bonsai}. Finally, a number of conferences and workshops have been established looking at the issue from different perspectives \citep{abdul2018trends}.                             

This explosion of XAI and IML research has been relatively ad-hoc, with many researchers attempting to make XAI systems specific to their area of AI - in other words, attempting to make existing AI algorithms/systems explainable. 
Miller et al. \cite{miller2017explainable} suggests that AI researchers typically build explainability from their perspective rather than the users'. 
There is a clear reason this is occurring. Current AI techniques are typically referred to as `weak' or `narrow' AI because they are designed to perform a single task. 
Therefore, when providing an explanation facility, the system is only required to explain how it performed within the context of that task. 
For example, to explain an image classifying Convolutional Neural Network (CNN), such as \cite{simonyan2013deep, zeiler2014visualizing, park2016attentive, goyal2016towards, wu2017interpretable, rajani:xai17, park2018multimodal}, a system only really needs to identify those parts of the image it focused on when classifying the photo or which parts of the neural network were activated. 
There is no need to justify its motivations or desires as it has none. 
These facilities are very useful for developers to verify that networks are focusing on the correct elements of an image to derive their decisions. 
Additionally, they can also provide assistance to novice users to build some trust on individual tasks. 
However, with the development of increasingly integrated intelligences, future systems will need to provide more general, broader and adaptable explanations that are aligned to the communication needs of the explainee (section \ref{subsection_example} will introduce a motivating example).  

Improved explanation of AI decision-making and behavior will also allow for the passive education of users. 
Awad et al. \cite{awad2018blaming} discusses the attribution of blame when an accident occurs in mixed (human and machine) control domains. 
Here it was found that people tended to blame the machine less than humans when an intervention was missed by the secondary driver --- e.g. where the AI did not take over from a human driver to avoid an accident. 
\citeauthor{awad2018blaming} \cite{awad2018blaming} surmises that this may be due to uncertainty on the machine's ability, while \citeauthor{Gray2012-GRAMPI} \cite{Gray2012-GRAMPI} suggests we need to understand the sort of mind dwelling in an AI to better attribute blame and causal responsibility. 
Therefore, improved explanation depth will provide better insights into the AI mind and revise people's expectations of AI capabilities, thereby, allowing people to make improved decisions about when they need to override control and when they can trust the AI system to behave correctly. 

In order to achieve these goals of trusted and socially acceptable systems, \citeauthor{miller2017explainable} \cite{miller2017explainable} emphasizes that AI systems should be modeled on philosophical, psychological and cognitive science models of human explanation. Such models have been researched for millennia and could be used to facilitate improved AI models for explanation \citep{miller2017explanationBook}. \citeauthor{miller2017explanationBook} \cite{miller2017explanationBook} provides three key areas of interest that should be of particular focus for good explanation: contrastive explanation, attribution theory and explanation selection. Furthermore, \citeauthor{hilton1996mental} \cite{hilton1996mental} suggests that acceptance of an explanation does not necessarily arise from correctness alone, but instead relies on pragmatic influences such as usefulness and relevance \citep{slugoski1993attribution,miller2017explainable}. \citeauthor{lombrozo2007simplicity} \cite{lombrozo2007simplicity} suggests that an explanation should rely on fewer causes (simple) that cover more events (general) and maintain consistency with peoples' prior knowledge (coherent) \cite{thagard1989explanatory}. This prior knowledge, \citeauthor{dazeley2008epistemological} \cite{dazeley2008epistemological} suggest, builds on the situation cognition view of knowledge, suggesting that people's knowledge is continually reinterpreted based on their current situation and hidden contexts from within their Merkwelt\footnote{Merkwelt - first used by \citeauthor{von1934} \cite{von1934} (translated into English \cite{von2013foray}) and later adapted to robotics \cite{brooks1986achieving, brooks1991intelligence}, refers to the complete set of environmental factors that have an affect on a species regardless of whether they are perceptible or not.}. 



This paper argues that in order for an agent to provide acceptable and trusted explanations it must continually determine an explainee's contextual position through an interactive process. During this process, typically referred to as a conversation, the agent progressively approaches the level of specificity required to lead the explainee towards the desired level of understanding \citep{miller2017explainable}. Furthermore, this paper will argue that AI cognition is structurally different to that of humans and that XAI systems need to go beyond cognitive interpretation alone and instead integrate technologies to achieve human-aligned conversational explanations. This paper's thesis will be achieved by providing definitions of explanation levels based on the cognitive process used to generate an AI system's decisions, and a mapping from these to the human social process that supports the three key areas of human explanation identified by \citep{miller2017explanationBook}. With explanations of AI decisions mapped to human models of explanation we aim to provide researchers and developers a theoretical basis to build better systems for trust and social acceptance. Throughout this paper we will discuss both current work and plausible approaches to address and integrate the different levels of explanation.

In summary, this paper makes four main contributions:
\begin{itemize} 
\item an argument, grounded in literature, for viewing XAI through a conversational lens (section 3);
\item a definition of levels of explanation with identified techniques that align with AI cognitive processes (section 4);
\item insights into the development of \textit{Broad eXplainable Artificial Intelligence (Broad-XAI)} (section 4); and,
\item a model for aligning AI explanation to human communication (Section 5).
\end{itemize}


\section{Motivating Example}
\label{subsection_example} 

Until recently, the majority of AI systems existed only on computer systems and our only interaction with these intelligences was when we chose to interact with them - such as accessing a website utilising an AI algorithm to recommend purchases. Likewise, cyber-physical AI-based systems tended to be limited to highly controlled environments, such as a manufacturing floor or sorting room. Such environments tend to be designed to exclude complexities caused by `contaminants' such as people and animals. Therefore, with no human interaction, there was little requirement for such systems to be concerned with providing complex explanations for their behaviour. In today's society, however, there is already a significant increase in the number of examples of integrated AI systems directly interacting with people, animals and other AI agents (e.g. personal assistance, healthcare, social robotics, navigation and security). In the coming years these systems will only increase in number and require more complex social interactions with users and bystanders.

A common example of a relatively integrated system already in use today is the autonomous vehicle. These systems use a range of sensors, cameras and communication systems to gather swathes of data about the environment that they cohabit. This allows them to perceive non-autonomous environmental features like road markings, trees and buildings, as well as details about other autonomous actors\footnote{Throughout this paper \textit{agent} will refer to the AI-based system making decisions that we are primarily concerned with providing an explanation, while \textit{actor} will refer to external autonomous beings, such as people, animals and AI-agents that are separate to the agent itself.}, such as cars, pedestrians and animals. These systems must also maintain a memory of what has happened previously, which can be used to construct a predictive model of other actors' potential future behaviour. Additionally, this memory, along with a range of logically induced sub-goals and learnt or user-set parameters, combine to provide the agent's internal motivations, such as its goals, beliefs and desires. In this paper we will refer to these internal motivations collectively as the agent's Merkwelt\footnote{von Uexk{\"u}ll \cite{von2013foray} suggested that a species' Merkwelt is determined by how it perceives the environment it inhabits, which in turn affects its behaviour and interpretation of the world. This interpretation is often extended indicating that each individual being has its own Merkwelt, such as \citeauthor{brooks1991intelligence}' \cite{brooks1991intelligence}. This suggests that a being's perception of the world alters its interpretation of the world, and therefore its internal motivations, which in turn potentially changes its behaviour.} ($\mathcal{M}$), extending \citeauthor{brooks1991intelligence}' \cite{brooks1991intelligence} suggestion that each autonomous AI-system has its own unique sensor suite, and therefore, will have its own Merkwelt affecting its behaviour. This extension incorporates the idea that internal motivations themselves can be affected by perceptions of the world, and therefore, will also form part of its Merkwelt. Finally, the autonomous vehicle makes a decision about any action, or non-action, that it must take, given the current environment and its Merkwelt, which may in turn affect the environment.

While such systems work well, as soon as they are placed in mixed environments the decisions made become increasingly complex. Importantly, these decisions become increasingly less obvious to the people with whom it is interacting, especially when something unexpected occurs. For example, in 2017 an Uber supervised self-driving car accidentally killed a pedestrian crossing the road \citep{knight2018uber} after apparently failing to take any avoidance measures. 
AI systems such as this should be able to explain to its user, insurance companies, authorities, and the general public what has occurred and the reasons why it made certain decisions, without having to rely on expert investigators to inspect log files. For instance, in such a situation a car's explanation may be: 
\begin{itemize} 
\item the system did not recognize a person pushing a bicycle; 
\item the system was motivated to remain in the right lane due to an approaching exit; 
\item the system believed that the person was giving way; 
\item the system believed that it had the right of way and that the person would expect the car to continue; or,
\item a combination of the above.

\end{itemize}

Additionally, the system's explanation itself and how it is generated should be verifiable to ensure it is not obfuscating the true reasons or attempting to intentionally deceive the explainee. Of course the reason may also be a more traditional fault such as mechanical/electrical failure, an incorrect user setting, or manufacturing fault. 
We are not suggesting why the Uber accident specifically occurred\footnote{Preliminary findings in the Uber incident released by \cite{NTSBReport}, revealed that the pedestrian was detected six seconds prior to impact and that 1.3 seconds prior, emergency braking was determined by the system to be required to mitigate a collision but emergency braking manoeuvres were not enabled to prevent the ``...potential for erratic vehicle behavior"\citep{NTSBReport}.}, but rather suggesting that such systems should be able to answer such questions themselves - as a matter of course. Furthermore, that the answer should be contextualised to the person receiving the explanation. For instance, the system's explanation to the user may be different to that provided to authorities and external parties. IEEE's P7001 Transparency of Autonomous Systems working group is developing standards around a suggested five different types of stakeholders: users, safety certifiers or agencies, accident investigators, lawyers and expert witnesses, and the wider public \citep{winfield2019ethical}. 

\begin{figure}
  \includegraphics[trim={2.7cm 4.9cm 2.6cm 4.0cm}, clip, width=\textwidth]{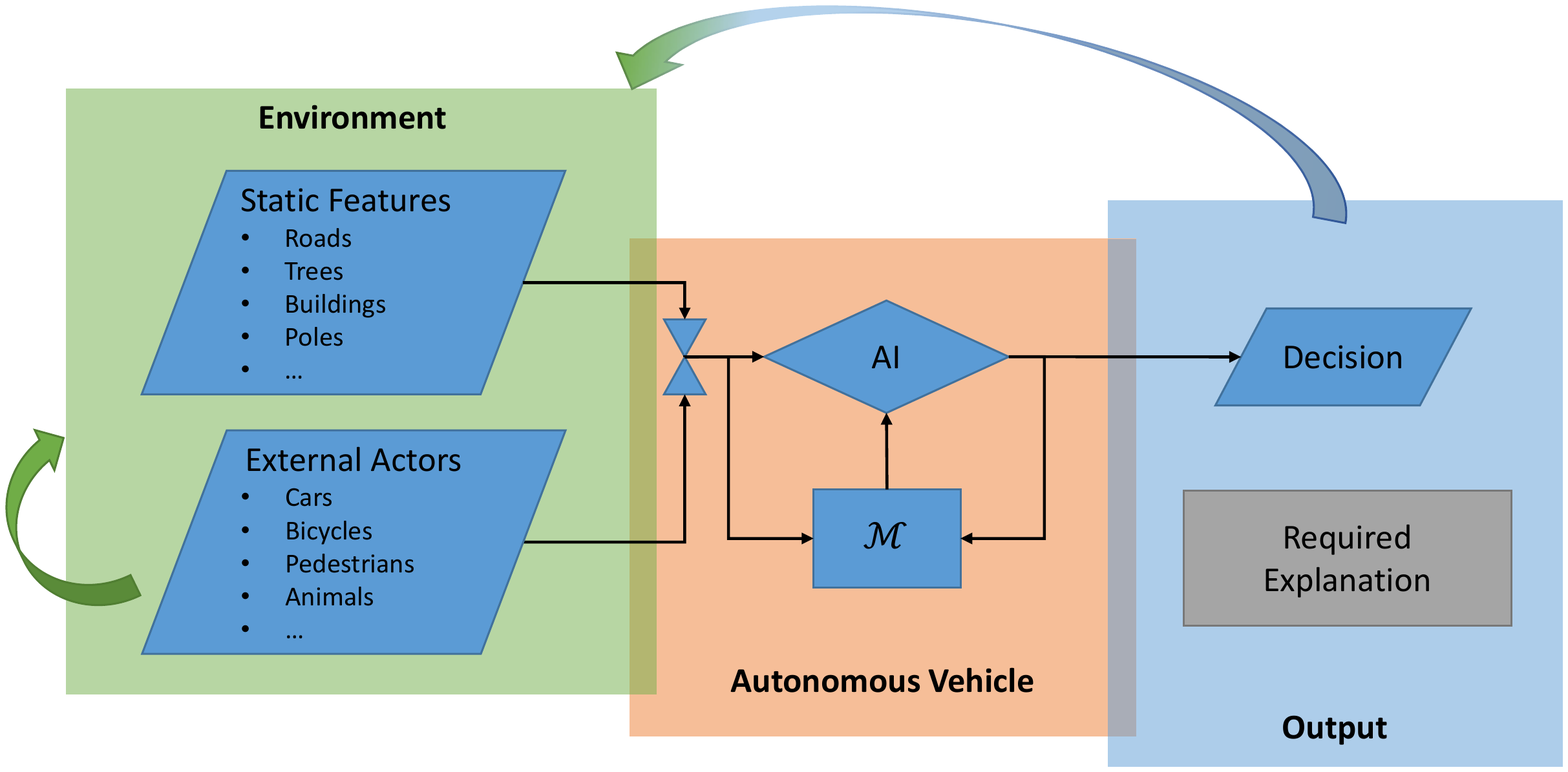} 
  \caption{\textbf{A Motivating Example} providing a diagrammatic representation of an autonomous self-driving car. The environment contains both non-autonomous features and external actors, whose behaviour may affect the environment, that need to be perceived by the agent. The agent combines the features perceived, illustrated using the collation flowchart symbol, with its internal disposition to form its Merkwelt ($\mathcal{M}$), which in many systems is derived from its memory of past events and either learnt or user set parameter settings. In this example the autonomous car makes a decision in the form of an action (or in-action), which may, or may not, affect the environment. It is the reasoning behind that action and its expected affect that requires explanation.
  }
  \label{fig:Auto_car}
\end{figure}

This example of an autonomous car and the need to explain its behaviour will be used as a motivating example throughout this paper. Figure \ref{fig:Auto_car} provides a simplified abstraction of an autonomous vehicle's interaction with its environment, which will be used to illustrate the concepts being discussed. Similarly, the suggested possible explanations provided above represent the range of explanation types that the various levels of explanation, introduced in section \ref{subsection_levels}, should be able to provide in a fully implemented Broad eXplainable Artificial Intelligent (Broad-XAI) approach. 

\section{Explanation as Conversation}
\label{subsection_Conversation} 

Arguably, current approaches to XAI are dominated by the interpretation of the decision-making process of the AI, rather than the provision of the reasoning about causation and expectations from those decisions and the communication of this to a generally lay audience. This is not surprising, as interpreting a decision is a relatively achievable and obvious first step. These approaches allow developers and researchers to understand an algorithm's operation in a similar way to a good quality debugging tool confirming our code operates correctly. 
However, such methods generally require a level of expert knowledge to fully understand, and there is relatively limited evidence they make significant inroads into the generally understood aims of XAI --- trust and understandability through the communication of decisions made to different audiences. This communication ultimately has a human on the receiving end, and therefore, we argue a full XAI-model must integrate the human into the explanation process. To design such a model requires an understanding of explanation and human communication, which this section will discuss. Subsequently, section \ref{subsection_levels} will break down the types of explanations into levels based on intentionality and identify current XAI research conducted at each level. Finally, section \ref{section_ConceptualModel} will discuss how such a system could be designed or implemented.




While explanation has been studied since Socratic times by philosophers and over the last fifty years by psychologists and cognitive scientists, what constitutes an explanation is still an active area of debate \citep{Woodward2017Explanation}. Therefore, which aspects of explanation should be taken up by XAI researchers is also an open question. Encyclopedia Britannica \cite{britannica2017Explanation} describes explanation as a set of statements that makes intelligible the existence or occurrence of an object, event, or state of affairs. According to \citeauthor{mayes2001theories} \citep{mayes2001theories}, ``Most people, philosophers included, think of explanation in terms of causation", but the term can also include deductive-nomological, deductive and statistical types of explanation \citep{britannica2017Explanation}. In Psychology, explanation of human behaviour tends to be focused around a person's social influences and internal beliefs and desires \citep{britannica2017Explanation,Mischel1963PsychologyExplanations}. \citeauthor{brown2006explaining} \cite{brown2006explaining} suggests that the roots of the term mean to `make plain' and that in standard English it most commonly means to `make known in detail'. Nevertheless, expatiate explanations --- provision of too much detail --- should also be avoided. For instance, \citeauthor{miller2017explainable} \cite{miller2017explainable} point out that explanation is a component of human conversation and that Grice's maxims \citep{grice1975logic}, for common and accepted rules of conversation, is that one should only say as much as is necessary for the explainee to understand. 

%

\citeauthor{hilton1996mental} \cite{hilton1996mental} presents a communication model of explanation in people as a social interaction where the explanation must be relevant. He suggest there are two stages to an explanation: diagnosis and the explanation itself. Diagnosis identifies causation, while the explanation is the social process to communicate the diagnosis. This aligns with \citeauthor{LOMBROZO2006464} \cite{LOMBROZO2006464} who notes that more generally an explanation can be considered both a process and the final product produced by that process. While \citeauthor{miller2017explanationBook} \citep{miller2017explanationBook} suggests that this process can be considered as requiring both a \textit{cognitive} and a \textit{social process}. The \textit{cognitive process} involves identifying a generalisable subset of causes, derived from attribution, along with possible counterfactual cases. Meanwhile the \textit{social process} involves an interaction between the explainer and explainee with the aim of providing the information required for the explainee to understand the explainer's \textit{proposed} causes of an event or outcome. Note, the explainer is only explaining \textit{proposed} causes, which are not necessarily the correct causes or even the causes the explainer believes are the correct causes \citep{LOMBROZO2006464, Wilkenfeld2015, miller2017explanationBook}. Hence, an explanation can intentionally be used to misinform the explainee based on the explainer's internal goal. 

\begin{figure}
  \includegraphics[trim={2.6cm 5.9cm 1.8cm 4.8cm}, clip, width=\textwidth]{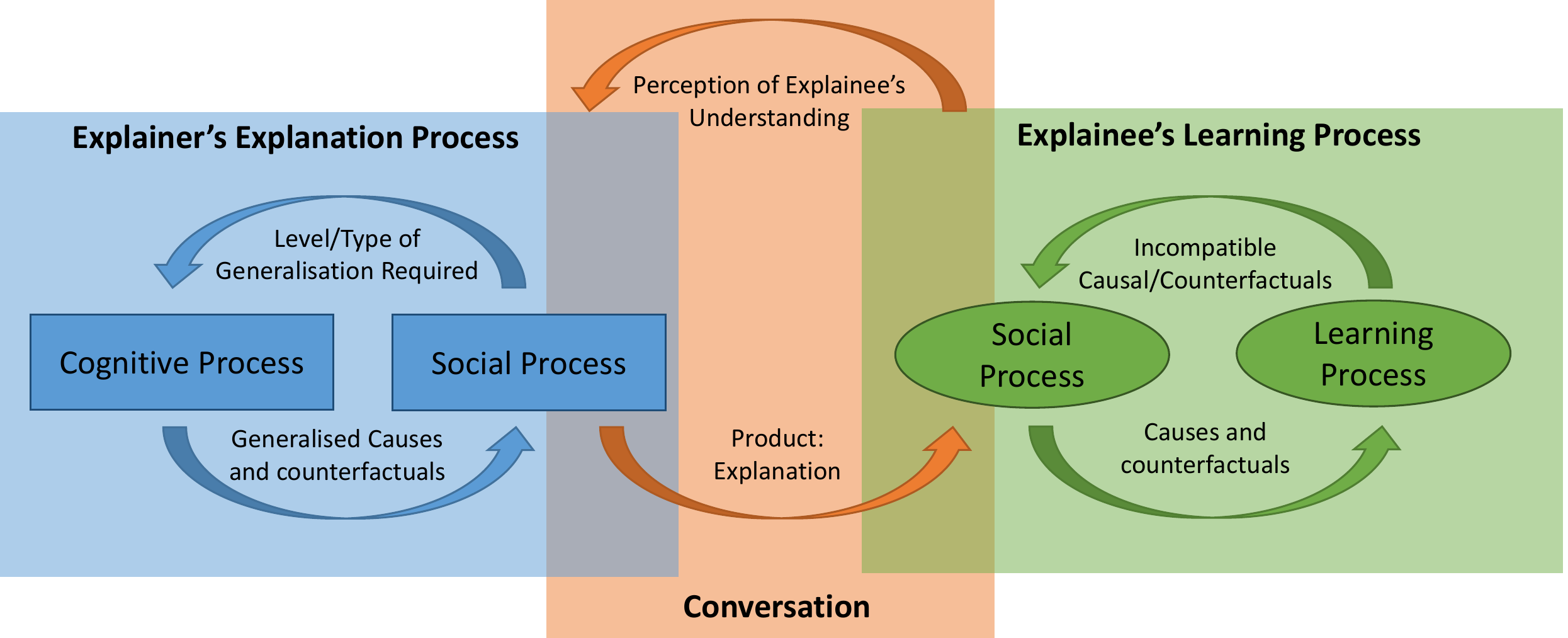} 
  \caption{\textbf{Explanation as a \textit{Conversational Process}}, illustrating the separation of the social and cognitive processes. The agent as a whole perceives the explainee's understanding and requirement for an explanation, of which the social process identifies the level and type of generalisation required from the cognitive process. The explainee must incorporate the explanation produced with its current world view and ask, explicitly or implicitly, for clarifications on any incompatibilities.}
  \label{fig:Conversation}
\end{figure}

\citeauthor{miller2017explanationBook}'s \cite{miller2017explanationBook} concept of separating the social from the cognitive process suggests that the social process forms what many authors refer to as the \textit{narrative-self} \cite{schechtman2011narrative, Harari2016} or \textit{remembered-self} \citep{salovey1993remembered, kahneman2011thinking}, while the cognitive process relates to the \textit{experience-self} \citep{kahneman2011thinking, Harari2016}. The latter, experience-self, refers to the fast, intuitive, unconscious process operating only in the present situation \cite{kahneman2011thinking}. In AI terms, this could be considered as the algorithm that is deciding on the immediate behaviour being followed by an agent. Hence, the experience-self is the cognitive decision-making process that AI researchers have been developing for 70 years. Whereas, the narrating-self is the conscious mind that creates explanations that support the internal goals of the individual \cite{Harari2016}. This explanation can be either to an explainee or to itself --- such as \textit{self-explanation} \cite{LOMBROZO2006464}. It is the narrating-self that directly participates in the social process that XAI researchers ultimately need to develop agent-human communication to facilitate understanding and trust in shared environments. \citeauthor{Harari2016} \cite{Harari2016} also suggests that the narrating-self only extracts a summary or generalisation of a person's experiences collected by the experiencing-self. Therefore, given all these suggestions that explanations must be a generalization and simplification of events and their causes, it is important that a model used by an AI system also incorporate the ability to generalise and simplify explanations that service the requirements of the explainee.

Yet, how does the explainer's cognitive process determine the degree and type of generalization required? While such a determination is itself a very difficult open research question being pursued by numerous researchers in fields such as culture modelling \citep{CultureModelling_Mascarenhas2016,CultureModelling_Hofstede2017}, user modelling \citep{UserModelling_Cawsey1993,UserModelling_Webb2001,PlayerModelling_Bakkes2012}, emotion recognition \cite{pal2019survey, mehta2019recent, rajan2019facial, chatterjee2019human, marechal2019survey, noroozi2018survey, supriya2019survey, li2018deep, salah2019video}, and human computer interaction \cite{biswas2012brief, biswas2010brief,nocentini2019survey,dutta2020human,truong2019social,ravichandar2016human}, we are interested in how the determination, once made, can be used to affect the explanation provided. We propose that this question implies that the cognitive and social processes form a cycle, as shown in Figure \ref{fig:Conversation}. The social process identifies the level and type of generalisation required to ensure the explainee's understanding and/or acceptance. The social process is, thus, informed by the explainee's reaction to its behavior or previously provided explanation (e.g. incorrect clarification statements, questions, observed body language or tone of voice). The identified level and type of explanation required is then forwarded to the cognitive process to identify the causes and/or counterfactuals that provide for the required level and type of explanation. Finally, the social process presents the causes and counterfactuals to the explainee in an appropriate form. 

This interaction between the explainer's and the explainee's social processes form the fundamental human process of communication, which in turn informs the cognitive process' level and types of causal and counterfactuals used. On the explainee's side the causes and counterfactuals are aligned to the explainee's currently accepted understanding of the world or more formally their objectivity illusion \citep{pronin2004objectivity}. If the presented causes and counterfactuals are not in conflict with their own then the explainee can accept them and incorporate them into their understanding of the world. Otherwise, the explainee can identify the causes or counterfactuals that are in conflict with their preconceived understanding. The explainee's social process can then: seek further, possibly deeper, explanations with the aim of resolving conflicts; find an alternative resolution of the conflicts (e.g. getting a second opinion); or, reject the explanation as incorrect. This process of accepting or rejecting an explanation aligns with common cognitive models of how people accept or reject arguments \cite{toulmin_2003}. 

As previously discussed, \citeauthor{miller2017explanationBook} \cite{miller2017explanationBook} suggests explanation systems should utilise existing human inspired models of explanation to inspire greater trust between machines and people. The difficulty with this assumption is that AI cognition differs significantly to human cognition \cite{Dickson2020difference} and extracting a human-like explanation from such algorithms is difficult. Whereas, by utilising this model of communication, where social and cognitive process are separated, allows us to isolate the the AI's cognitive process from how the explanation is expressed to a person. Importantly, this also means, instead of classifying explanation systems on the \textit{final product} (the explanation), or the \textit{cognitive process} (decision-making algorithm) used; we can classify them based on the \textit{social process} required to address the explainee's requirements. This model is put forward with the aim of aiding developers and researchers to integrate the required systems to provide the socially required explanations.


\section{Levels of Explanation}
\label{subsection_levels} 

The aim of eXplainable Artificial Intelligence (XAI) is to provide explanations for decisions/conclusions made by AI systems that people can understand and accept. Yet without a strong definition of what an explanation is in human society, means that XAI has also been unable to provide a consistent definition for practical applications. For instance, Gunning from DARPA's XAI project simply describes XAI's aim as ``...finding a way to put labels on the concepts inside a deep neural net'', as quoted by \citep{kuang2017nytimes}. Other XAI researchers, such as Rosenthal \cite{rosenthal2016why} from the SEI Emerging Technology Center of Carnegie Mellon University, state that their goal is to translate robot action, written in code, to English. While both of these descriptions are probably simplified for lay audiences, they do represent a common theme in the field at the moment - that explanation primarily involves the provision of a literal translation of individual decision points.

In \textit{`Explaining Decisions Made With AI'} (2020) \cite{Turing2020XAI}, a co-badged report by the Information Commissioner's Office (IOC) and The Alan Turing Institute (ATI), XAI is described as consisting of both technical (extraction of information) and non-technical (method of communication) considerations. Technical considerations are categorised as either \textit{local} or \textit{global}, where a local explanation provides an interpretation of individual predictions, while global explanations describe a system's behaviour across all outputs. While global explanations are crucial for understanding the algorithm's performance holistically, this is unlikely to be of significant importance to an individual using the system day-to-day. \citeauthor{kazim2020explaining} \cite{kazim2020explaining} suggests that local explanations should be prioritised as they focus on the individual that may be affected by the decision. Generally, we find this binary distinction simplistic and restrictive. Hence, while this paper focuses on the provision of individual decisions through local explanation, we recognise that these local explanations can be supplemented with global-like explanations of sub-components that affect the outcomes of the individual decisions.

To address local explanations the IOC and ATI report \cite{Turing2020XAI} advocates combining and integrating explanation strategies --- what we are referring to as \textit{Broad-XAI}. Here they describe three levels of explanation: Visualising how the model works; Understanding the role of variables and variable interactions; Understand how the behaviours or circumstances that influence an AI-assisted decision would need to be changed to change that decision \cite{Turing2020XAI}. The issue with these levels is they are highly focused on the technical aspects with the non-technical communication and human needs being considered separately. Importantly, these levels facilitate a focus on interpretive XAI and do not consider the social and cultural context of the AI's decision or how to communicate this to the explainee. As discussed in the previous section, full XAI should be considered as containing two closely interacting processes:

\begin{itemize} 
\item social process - interacting with people, animals and/or other agents; and,
\item cognitive process - identifying appropriately generalized causes and counterfactuals.
\end{itemize}

In this paper we argue that explaining individual decision points in the context of the inputs at that time, while important in some cases, is only one part of the explanation problem. These explanations do not attempt to understand the social context of the explainee and do not usually find a generalisation over the causes or couterfactuals. As AI systems increasingly become integrated into our everyday society, simply explaining a single decision point without the historical origins of that action \citep{Dennett1987-DENTIS} or its social/cultural context will not carry sufficient meaning. We suggest that the agent may also need to describe what it was attempting to achieve, or what other environmental factors influenced (or did not influence) that decision.

\begin{figure}
  \centering
  \includegraphics[trim={6.1cm 2.5cm 5.3cm 2.7cm}, clip, width=24pc]{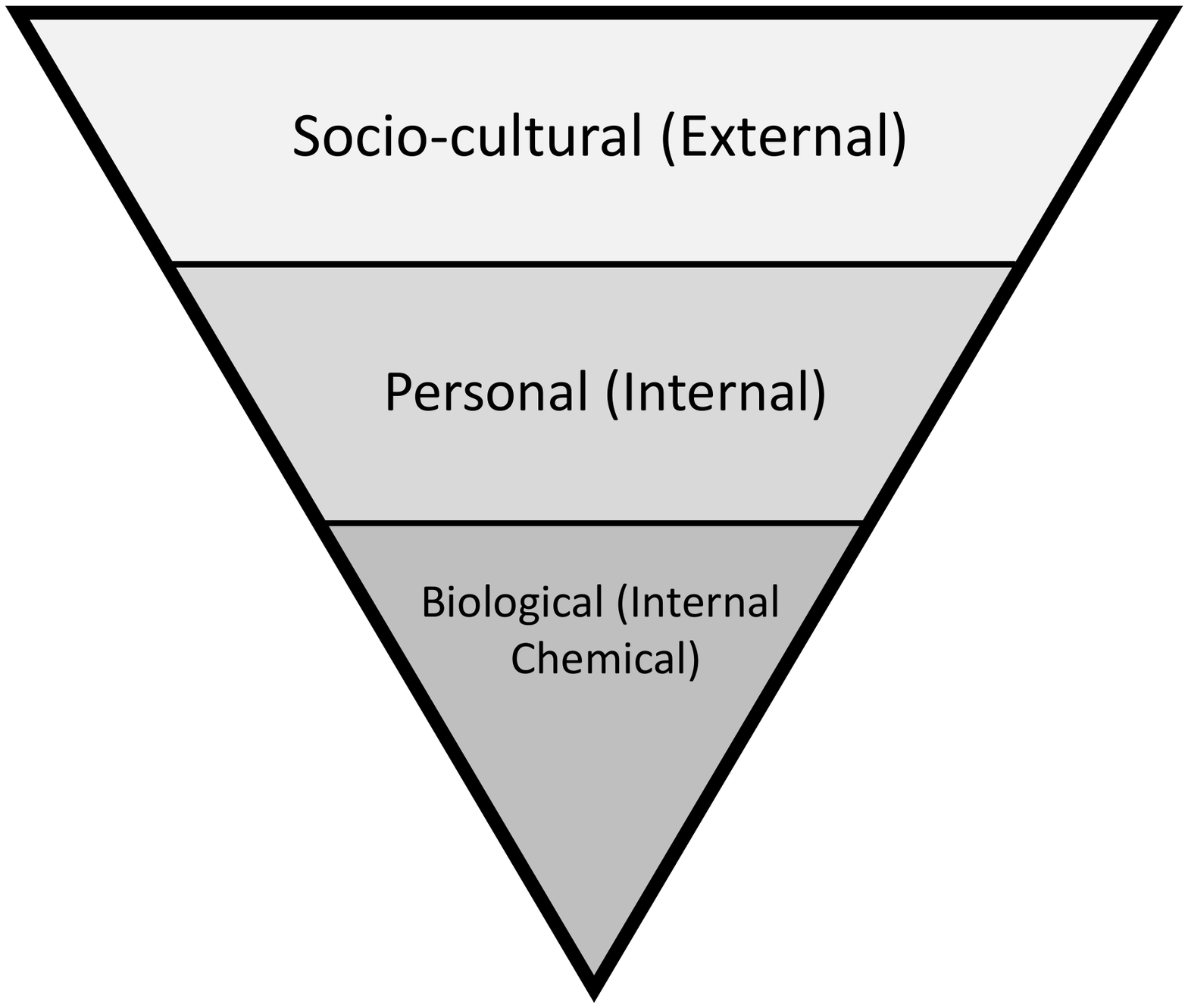}
  \caption{\textbf{Psychological Levels of Explanation} \cite{stangor2010IntroductionPsychology}, illustrating the reductionist model generally used to explain human behaviour.}
  \label{fig:psychologyLevels}
\end{figure}

Miller \cite{miller2017explanationBook} suggested that before XAI can begin to produce broader explanations, we need to understand the different levels of explanation and how they might be generated from the cognitive processes of an AI system. Since Aristotle a number of philosophers have identified different models or levels to categorize various types of human explanation \citep{Hankinson2001}. For instance, Aristotle describes \textit{four causes} of explanation: Material, Form, Agent and End. Other authors have extended this model to include an actor's intentions \cite{Dennett1987-DENTIS, Kass1987}. While these provide the material form or structure of actual explanations, they provide minimal insight into how an explanation is cognitively created, which is the problem faced by AI researchers. Cognitive Science aims to determine the cognitive process of mind, and therefore, offers more potential. For instance, Marr's Tri-Level Hypothesis \cite{man1982computational} identifies three Levels of Analysis: computational, algorithmic and implementational. These were later extended by \citeauthor{poggio2012levels} \cite{poggio2012levels} to include learning and evolution. However, these are focused on categorising the computational approach in \textit{generating the decision} rather than the \textit{generation of the explanation}. 

In Psychology, any standard textbook, such as \citeauthor{stangor2010IntroductionPsychology} \cite{stangor2010IntroductionPsychology}, identifies three levels for explaining human behaviour based on a reductionist model (see Figure \ref{fig:psychologyLevels}): Social and Cultural (External Influences); Personal (Internal psychological)\footnote{Sometimes this is broken into two separate levels of Personality and Basic Process (learning, memory)}; and, Biological (Internal chemical influences). This model suggests that explanations are provided from the top-down, focusing first on the social and cultural explanations and drilling down to deeper levels where required. The advantage of this model is that it successfully divides the cognitive processes into the different influences on a person's behaviour, which could work well in identifying the cognitive influences on an AI agent when making a decision. The difficulty here is that AI systems do not use the same socially-focused cognitive process as people to generate their decisions. Instead their decision-making is primarily experientially (data) driven. 

\begin{figure}
  \centering
  \includegraphics[trim={5.2cm 2.9cm 6.2cm 2.4cm}, clip, width=24pc] {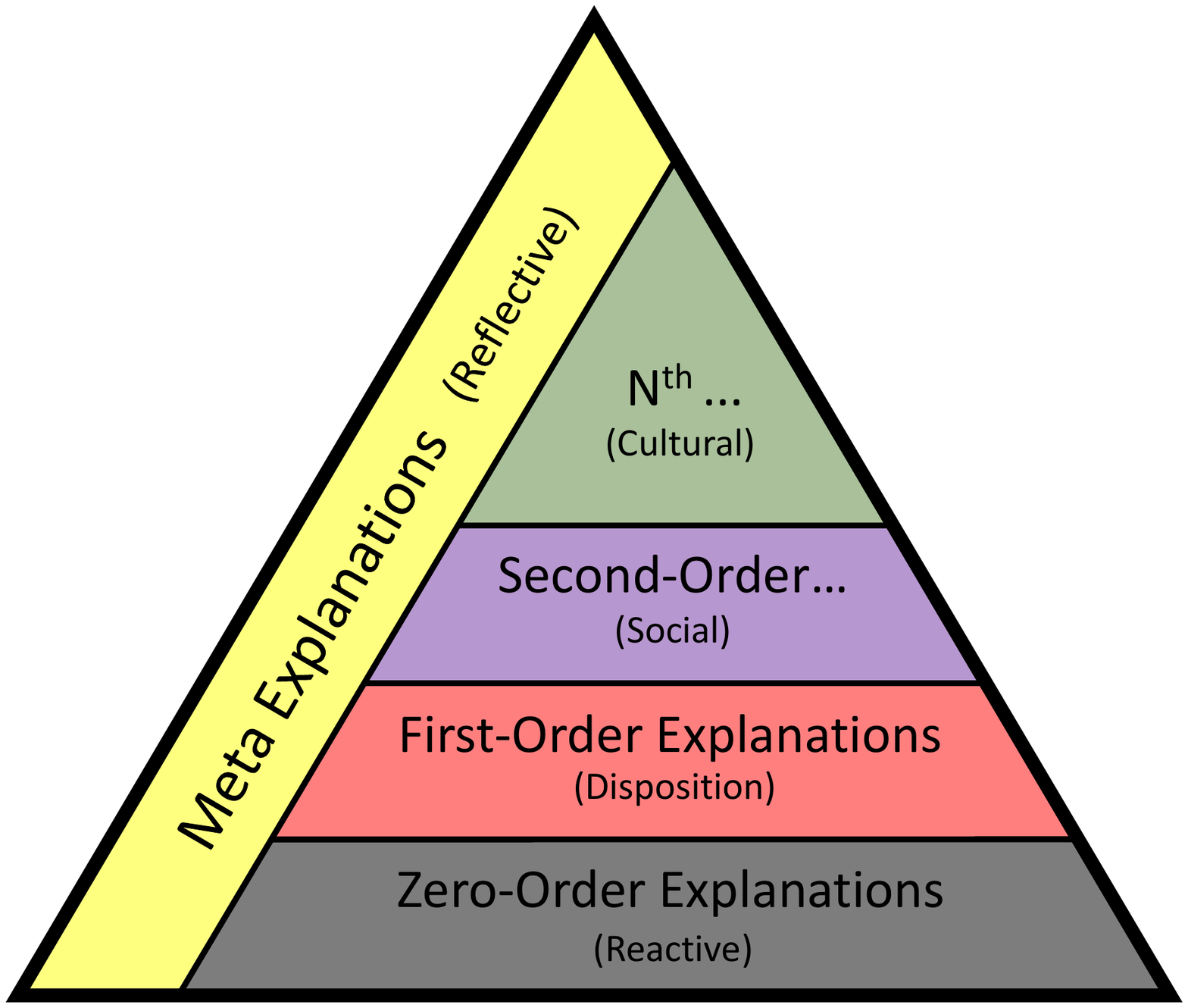}
  \caption{\textbf{Levels of Explanation for XAI}, providing a bottom-up constructivist model for explaining AI agent behaviour. This model is adapted from Animal Cognitive Ethology's levels of intentionality \citep{griffin1976animalawareness, dorothy1990monkeys}}.
  \label{fig:Levels}
\end{figure}

Therefore, instead of this reductionist approach, we argue that a model for different levels of explanation for AI requires a bottom-up constructivist approach that aligns with the types of cognitive processes used by AI technology. This goes against Miller's \citep{miller2017explanationBook} arguments that levels should be based on the above theories. We suggest instead that it is more useful to have a model that aligns with AI's cognitive process that also allows for the generation of explanations suitable for the social process. To achieve this we propose to adapt work from the field of Animal Cognitive Ethology on Intentionality\footnote{Animal Ethology explain animal behaviour through levels of intentionality. In XAI we are attempting to explain AI intentionality.} behind observed behaviour \citep{griffin1976animalawareness, dorothy1990monkeys}. Animal Cognitive Ethology has been designed to provide explanations of animals' behaviour based on a spectrum of intentional levels. We argue that this spectrum provides greater alignment with current AI approaches to cognition while still providing the explanation versatility required for social agents. Based on this approach we will define four levels of explanation plus a Meta-explanation --- as illustrated in Figure \ref{fig:Levels} and discussed extensively in the following sub sections.

\subsection{Zero-order (Reactive) Explanations}
\label{subsection_zero} 

As discussed earlier a significant amount of current XAI work focuses on the interpretation of a single decision point based on data provided to generate that decision. Sometimes referred to as local explanations \cite{mueller2019explanation}, the aim of these systems is to identify why the system provided a particular decision/conclusion/value/classification based on the data provided for that instance. In this paper, we introduce the term \textit{Zero-order explanations} to formally categorize these types of explanations, but will generally refer to them by the more descriptive name of \textit{Reactive explanations}: 





\bigskip
\begin{addmargin}[3em]{2em}
\textbf{Zero-order (Reactive) Explanation:} \textit{is an explanation of an agent's reaction to immediately perceived inputs.}
\end{addmargin}
\bigskip

\begin{figure}
  \centering
  \includegraphics[trim={3.3cm 4.0cm 2.2cm 3.4cm}, clip, width=\textwidth] {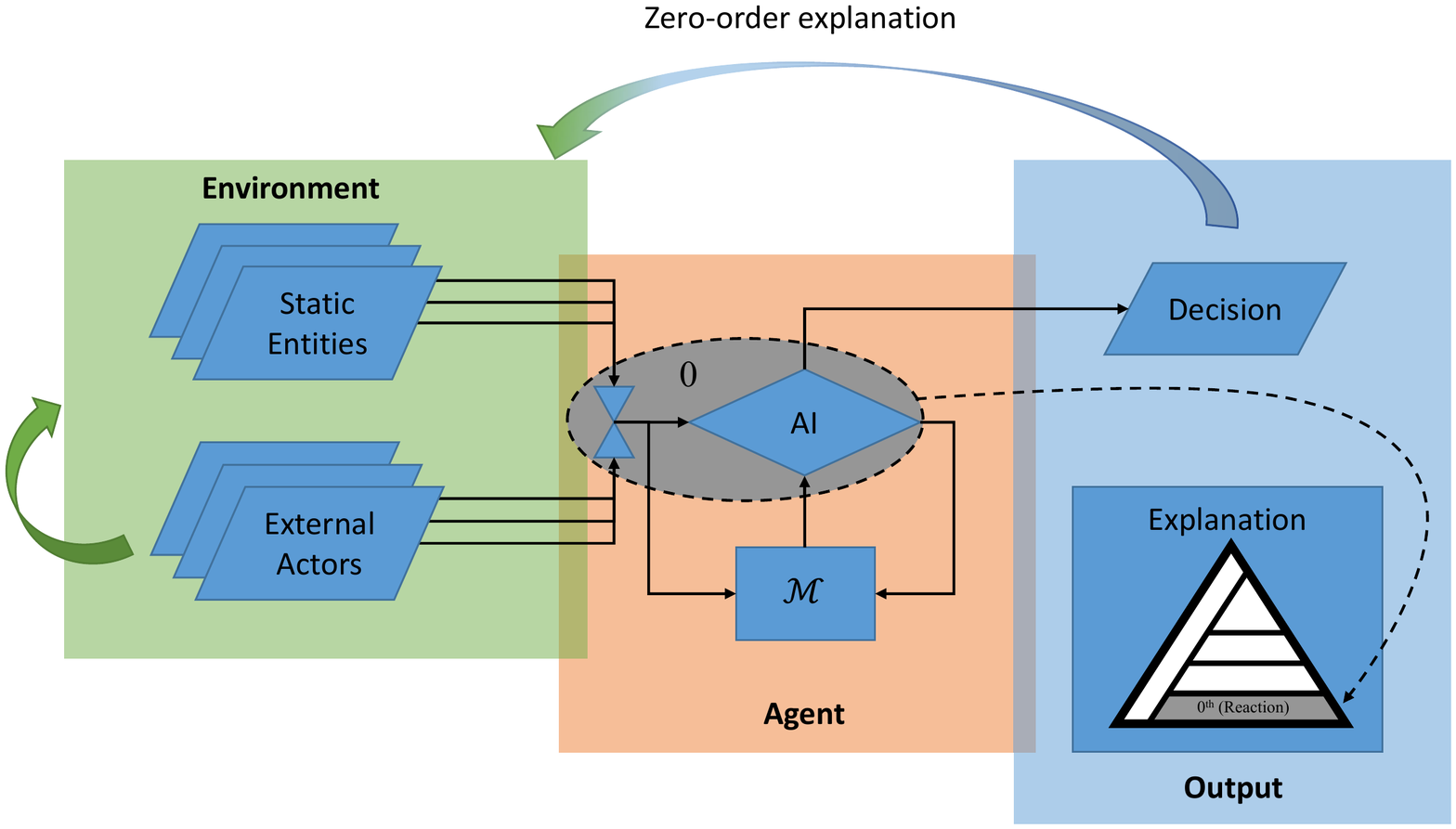}
  \caption{\textbf{Zero-order (Reactive) Explanation} shown diagrammatically, where the grey oval, $0$, indicates the focus area of Reactive explanations around the AI agent's decision based on the interpretation of the input.}
  \label{fig:ZeroOrder}
\end{figure}

This definition is designed to encompass all explanations that are made to describe a decision that is a reaction to a single set of presented data or its current environmental state. In Ethology this is based on the idea that animals may have no (\textit{zero}) intentions when simply automatically reacting to a situation \cite{dorothy1990monkeys}, and thus, an explanation of their behaviour is based solely on their immediate environment. For example, a vervet monkey raises an alarm when it perceives a danger such as a snake in the grass. If the monkey is behaving with Zero-order intentionality then it has no motives behind raising the alarm, it simply automatically raises an alarm in the presence of danger \cite{dorothy1990monkeys}. Figure \ref{fig:ZeroOrder} illustrates the area of focus for such explanations in a typical autonomous agent. This is a generalisation of the autonomous car example in Figure \ref{fig:Auto_car}, where the
Reactive explanation for an agent is focused on the area indicated by the grey oval, $0$, representing the process of extracting features perceived and deriving a decision. 

In AI, the vast majority of current approaches are known to be purely reactive non-intentional systems, and therefore, any explanation provided by such a system can only ever be at a Zero-order level. For instance, a convolutional neural network is constructed with a number of stages, each with one or more layers of neurons. The network provides a function that maps inputs to an output representing the system's decision. By analysing the low levels you can provide an explanation of the features identified in the input. By analysing the degree to which particular neurons throughout the network fire you can provide an explanation of how those features influenced the decision. In our motivating example, this level of explanation allows the agent to answer questions such as: `what input features were the most important when making a decision?'; `what objects were observed in the environment?'; or, `what features of the state made it decide to turn left?'. This level of explanation is also appropriate to answer many counterfactual style questions, such as Doshi-Velez et al.'s \cite{doshi2017accountability} questions: `Would changing a certain factor have changed the decision?'; or, `Why did two similar-looking cases get different decisions, or vice versa?'.

Reactive explanations, while the lowest level, are possibly one of the most valuable. Figure \ref{fig:Levels} uses a pyramid structure to illustrate that, while Reactive explanations are at the bottom, they are the foundation to all other explanations that are built on top of this level. It is also the area currently most researched, and is commonly referred to as Interpretable Machine Learning (IML) in the literature \cite{guidotti2018survey, doshi2017towards}. It is imperative in any AI implementation that we understand how a system directly interprets its immediate surroundings and inputs. These explanations allow us to ensure that the correct inputs are being focused on and correctly processed. For instance, there are numerous examples throughout AI development of systems that appear to have worked well, but after closer investigation, where found to have derived the correct answer from features provided inadvertently by the engineers. An example, is the urban legend of a tank classification system that correctly classified whether tanks were in an image. However, upon investigation it was found the system classified images based on the direction of the shadows cast by trees on the ground. This was reportedly caused by inadvertently photographing the tanks in the morning and the photos without tanks being taken in the afternoon \citep{whitby2009artificial}. An explanation system that identifies those areas of an image that were the main focus of the AI's attention would have quickly revealed to a human observer that it was not looking at the tanks at all. These interpretation based explanations provide researchers and developers the ability to debug their utilisation of the algorithm and identify how they may be able to improve performance.

There are countless examples of XAI research that can be categorized as Reactive explanation. Guidotti et al. \cite{guidotti2018survey}, after surveying the literature, identified that IML based explanations of black-box Machine Learning systems fall into \textit{model}, \textit{outcome} or \textit{model inspection} problem types. Guidotti et al. \cite{guidotti2018survey} suggest that people will either be presented with a transparent \textit{model} of the black-box that mimics the behaviour, such as a set of rules allowing the user an interpretation of how the system will behave for individual cases \cite{huysmans2011empirical}. Alternatively, the system will provide an interpretation of a single instance's \textit{output}, such as an image, text or graph, illustrating how an individual data instance was processed \cite{fong2017interpretable, xu2015show}. Finally, the \textit{model inspection} problem provides either a model or output representation of a sub-component of the black-box, such as sensitivity to an attribute change or neural network activation levels \cite{zintgraf2017visualizing, SundararajanTY17}. 


\subsection{First-order (Disposition) Explanations}
\label{subsection_first} 

While Reactive explanations are the foundation of any explanation system, there is clearly a requirement for more meaningful explanations - especially as AI systems become more advanced. As \citeauthor{tegmark2017life} \cite{tegmark2017life} suggests, we need to understand an agent's goals, objectives, beliefs, emotions and memory of past events and how these affect its reaction to a set of inputs. For instance, methods such as Belief, Desire, Intention (BDI) agents and Reinforcement Learning are already designed as goal-oriented or intentional AI systems. While these systems are still primarily reactive (e.g. given the state of the Go board the agent makes a particular move), the way they react is based on predefined or sometimes even learnt objectives. The only way people can understand why an agent has reacted in a particular way is if the system provides its context based on its current goal or other factors driving its behaviour. \citeauthor{langley2017explainable} \cite{langley2017explainable} describes this as \textit{explainable agency} and focuses on the idea that many agents are goal-directed. In particular, agency can be considered as the capacity of an agent to act independently - making its own free choices. In terms of explanation levels we expand on the concept of agency to be the agent's internal disposition and define the term \textit{First-order explanations} to formally categorize these types of explanations. and refer to them descriptively with the name \textit{Disposition explanations}: 



\bigskip
\begin{addmargin}[4em]{3em} 
\textbf{First-order (Disposition) Explanation:} is an explanation of an agent's underlying internal disposition towards the environment and other actors that motivated a particular decision.
\end{addmargin}
\bigskip

This definition builds on Reactive explanation by also considering the agent's current internal disposition when making its decision, such as its \textit{belief}\footnote{\textbf{Belief:} an agent has either a predefined or a learnt `truth' about the environment in which it is acting. For example, an agent may have a \textit{memory} or \textit{prediction} of an object's location even though it can no longer be directly perceived.} and/or \textit{desires}\footnote{\textbf{Desire:} an agent has either a predefined or learnt aim to acquire or achieve one or more things. For example, an agent may desire a tidy room. In AI research desire can incorporate, and is often used interchangeably with, the term: objective and/or goal. The precise word used is often dependent on the specific subfield of AI being implemented.}. A Disposition explanation may, and usually will, still draw on the foundational (reactive) explanation, but will also incorporate details of the agent's current internal disposition and how it influenced its reaction. In Ethology, this is based on the idea that animals' behaviour can be based on its current intentions that come from some form of internal belief and/or desire. For example, a vervet monkey's belief that there is a risk of a nearby predator or the desire for sweet fruit \citep{dorothy1990monkeys}. In these situations an explanation would include the internal drive, such as the desire for sweet fruit (Disposition explanation), with the Reactive explanation of perceiving a rich colour through the trees that could be fruit. In such a situation a purely Reactive explanation would simply indicate that it saw fruit, which without a Disposition explanation would provide no meaningful justification for its reaction to move towards the rich colour. 

\begin{figure}
  \centering
  \includegraphics[trim={3.3cm 4.0cm 2.2cm 3.4cm}, clip, width=\textwidth] {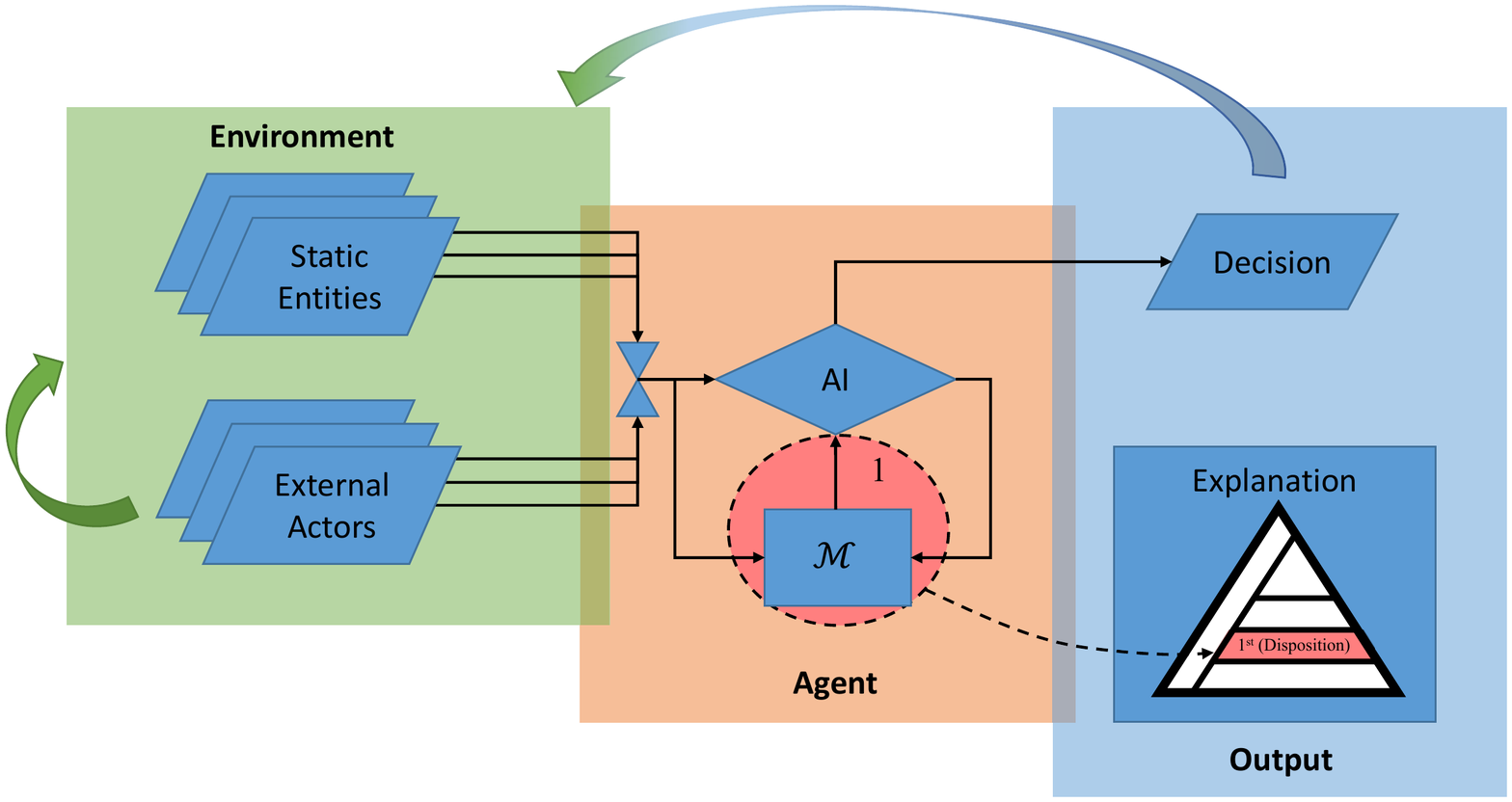}
  \caption{\textbf{First-order (Disposition) Explanation} shown diagrammatically, where the red oval, $1$, indicates the focus area of Disposition explanations around the AI agent's decision, based on the interpretation of the agent's Merkwelt, $\mathcal{M}$, such as its beliefs, desires and/or memories.}
  \label{fig:FirstOrder}
\end{figure}

Represented by the red oval, $1$, in Figure \ref{fig:FirstOrder} illustrating the area of focus, Disposition explanations allow agents to answer questions such as: `why did you want to be in that lane?'; `why did you want to clean the room?'; or counterfactuals like `why didn't you wish to play the rook to c7?'. It is important to note the difference between the questions asked that resulted in a Reactive versus a Disposition explanation. Reactive explanations are provided when the question asked about how the inputs to the agent contributed to the decision. These allow us to verify its perception and response to its inputs. A Disposition explanation, on the other hand, is provided when the question asks about the agent's internal motivations that caused it to react to the inputs in a particular way. Many current predictive AI systems (such as Decision Trees, Random Forests, Convolutional Neural Networks, Naive Bayesian Classifiers, SVMs, Regressions) have no inherent internal motivations other than their programmed and global need to make a prediction based on a set of inputs. Hence, any explanation of such a system's motivations would equate to the incorporation of a global explanation and would remain relatively meaningless to ask such a system a question requiring a Disposition explanation. 
    

On the other hand, there are also a number of AI approaches that are designed to have, or can utilise, some notion of belief and/or desire when determining a decision, such as Reinforcement Learning, Bayesian Optimization, Dynamic Programming, BDI agents, Evolutionary algorithms, Bayesian Networks, and even search algorithms like A*. For instance, a Reinforcement Learning agent \citep{Sutton2018book} is designed to search for an optimal solution to achieving a prescribed goal, while a BDI agent \citep{rao1995bdi} is engineered to achieve a desired outcome based on predefined beliefs and desires. However, many of these approaches do not have any inbuilt knowledge or control over these beliefs or desires, nor do they require this information as they generally only have a single goal that is usually defined by the user/engineer. It is, therefore, not particularly useful to provide a Disposition explanation in these simple systems.

Whereas, consider a machine learning approach that incorporates various ethical and safety operations that modify behaviour from a primary objective, such as suggested by \citep{vamplew2018human} and implemented by \cite{vamplew2020potential}. In such a system these ethical and safety objectives act as internal dispositions of the agent. If a user needs to understand the algorithm's behaviour then simply explaining the interpretation of the immediate environment is insufficient and details of the current objective, ethical consideration, or safety protocol being followed is key.

These areas of study are relatively recent and there has been very limited research in the incorporation of explanation systems into these approaches. There is currently no distinct subfield of XAI research that could be classed as encompassing this level of explanation. Although recently, \citeauthor{anjomshoae2019explainable} \cite{anjomshoae2019explainable} suggested the term goal-driven XAI encompassing part of this concept. The most developed Disposition explanation work is based on BDI agents \cite{rao1995bdi}. BDI agents could be considered as an extension to earlier knowledge engineering approaches to AI, in that they are typically engineered solutions. Like knowledge based systems, BDI agents are grey-box systems\footnote{A grey-box algorithm combines theoretical or engineered structure with live data \cite{kroll2000grey}. A grey-box algorithm can often be directly interpreted to provide an explanation.}, and therefore, well suited to the provision of an explanation. The focus of BDI agents is the modeling an agent's beliefs and desires and how these change over time due to their social interactions. Therefore, they are particularly well suited to providing Disposition explanations. For instance, Harbers et al. \cite{harbers2011explanation, harbers2011explaining, harbers2010design} developed agents that recorded behaviour-logs of their past mental states and corresponding actions. These approaches could provide explanations for intentional behaviours such as it opened the door `because it believed someone was outside' \cite{harbers2010design}. In Reinforcement Learning and Markov Decision Processes (MDPs) more generally, initial work in goal directed explanations can be seen in \cite{madumal2019explainable, cruz2019XRL, khan2009minimal}, where causal models or predictions about the outcomes are used to explain an agent's behaviour.

The issue with these approaches are that Disposition explanations are not exclusively derived from an agent's goals, objectives and beliefs, but may also be based on an agent's emotional state, and/or its memory of past events. For instance, many learning systems rely, not only on their current inputs, but also on a memory of past events. Disposition explanations should be able to provide details of how these elements of their Merkwelt influence their decisions. For instance, the family of Recurrent Neural Networks (RNNs) such as, Long Short-Term Memory (LSTMs) and Gated Recurrent Units (GRUs), combine current inputs with a directly propagated memory of past events to determine its outputs. Approaches have been developed to explain or provide interpretations of their decisions based on both the input and the carried forward memories, such as \citeauthor{arras2017explaining} \cite{arras2017explaining} that produces heat maps during sentiment analysis, and \citeauthor{bharadhwaj2018explanations} \cite{bharadhwaj2018explanations} that explains temporal dependencies. Similarly in the recently coined field of Emotion-aware XAI (EXAI) an agent includes its own emotional state to explain its behaviour \cite{kaptein2017role}. Human actions are frequently expressed on the basis of one or more emotions behind an action \cite{rorty1978explaining}. EXAI agents first have inbuilt cognitive processes for formulating an emotion that is expressed through an explanation formalism \cite{o1994explaining, kaptein2017role}. Memory and emotion-aware explanations cannot adequately explain an action by themselves and, as supported by \citeauthor{kaptein2017role} \cite{kaptein2017role}, need to be combined with the agent's beliefs, desires and goals, along with its direct interpretation of the current state to explain its decision.

For an agent to include an explanation that incorporates both Reactive and Disposition explanations, such as EXAI approaches \cite{kaptein2017role}, the algorithm being explained will generally be an amalgam of two or possibly more base level algorithms. Currently, research into such systems may still be regarded as IML. However, we suggest that this is misleading. Such systems need to merge multiple interpretations from different machine learning algorithms into a single coherent interpretation for an agent's behaviour. Instead, we propose that such system should be regarded as \textit{Broad eXplainable Artificial Intelligence (Broad-XAI)}. Broad-XAI research focuses on combining interpretations that are extracted from multiple base algorithms. For example, Deep Reinforcement Learning \cite{li2017deep, arulkumaran2017brief, duan2016benchmarking, hossain2019comprehensive} is a class of algorithms that essentially utilise a combination of two distinct algorithms Reinforcement Learning and Deep Neural Networks. To build a system that incorporates both Disposition and Reactive explanations, you must merge both the interpretation of the agent's goal from the RL transition model (Disposition explanation), with the interpretation of the current state from Deep Neural Network performing the function approximation or feature extraction (Reactive explanation). Currently, most explanations of Deep-RL only explain the reactive parts of the algorithm \cite{madumal2019explainable, lee2018modular, hendricks2016generating} and would not be regarded as Broad-XAI. This is primarily because the approaches have a preset goals or objectives. As discussed earlier approaches where the goal is learnt or derived such as Deep Multiobjective Reinforcement Learning \cite{nguyen2018multi, abels2018dynamic, mossalam2016multi, ferreira2018multiobjective} offer an interesting platform for Broad-XAI.

\subsection{Second-order (Social) Explanations} 
\label{subsection_second} 

As research moves closer to Artificial General Intelligence, simply understanding an agent's internal belief and/or desire is insufficient. Animal Ethology recognised that some animal behaviour may indicate an awareness of, or at least an interpretation of, other animals' internal beliefs and desires \citep{dorothy1990monkeys}. For instance, a vervet monkey in a zoo sees the keeper carrying a bucket into the enclosure and concludes they intend to feed it. In Animal Ethology there is disagreement as to whether the animal is aware of the keeper's intentions or is simply predicting future behaviour. Regardless, this and higher levels of reasoning about mental-states are recognised in humans as metacognitive processes. This could be based on reasoned mental states as expressed in the \textit{Theory of Mind} \cite{leslie1987pretense}; or, through imagined mental states as expressed by the concept of \textit{mentalisation} \cite{wimmer1983beliefs} --- colloquially often referred to as empathy, emotional understanding, attribution, mind-mindedness and self-awareness. Mentalisation can be considered as the ability to see ourselves as others see us, and others as they see themselves \citep{holmes2008mentalisation}; or, more formally, it is the recursive understanding of mental states \citep{lewis2017higher}. Both this and the next level of explanation in this paper are based on deeper recursive levels of mentalisation. Current evidence suggests that mentalisation may be unique to humans \citep{saxe2006uniquely, tomasello1997primate} and can be regarded as core to social cognition in humans and is thought to develop in children around the age of 4 or 5 years \citep{lewis2017higher}. 

As AI agents become increasingly integrated within our society, it is clear that in order to function and be accepted, they will require the ability to model other actors' (people, animals and agents) behaviour \citep{AGI_Adams2012,AGI_Goertzel2014,AGI_Muller2016, rabinowitz2018machine}. For example, in order for the Uber self-driving car, discussed earlier, to function the system requires a model of behaviour for other actors in the environment --- such as the pedestrian with a bicycle. However, if an agent simply models behaviour then when providing an explanation it can only discuss the expected behaviour, but not provide any reasoning about \textit{why} it expected that behaviour. In order for an agent to describe another actor's intentions, it will also need the model to predict the actor's internal state and likely memory. This model provides a belief of other actors' mental states and allows for the prediction of both their behaviour and their likely motivations for that behaviour. Agents that are provided with or learn such a model will also require the ability to explain decisions made based on this model. In this paper, we define a \textit{Second-order explanation} to formally categorize these types of explanations, and will generally be referred to by the more descriptive name of \textit{Social explanations}:

\bigskip
\begin{addmargin}[4em]{3em}
\textbf{Second-order (Social) Explanation:} is an explanation of a decision based on an awareness or belief of its own or other actors' mental states.
\end{addmargin}
\bigskip

\begin{figure}
  \centering
  \includegraphics[trim={2.9cm 4.1cm 2.2cm 3.2cm}, clip, width=\textwidth] {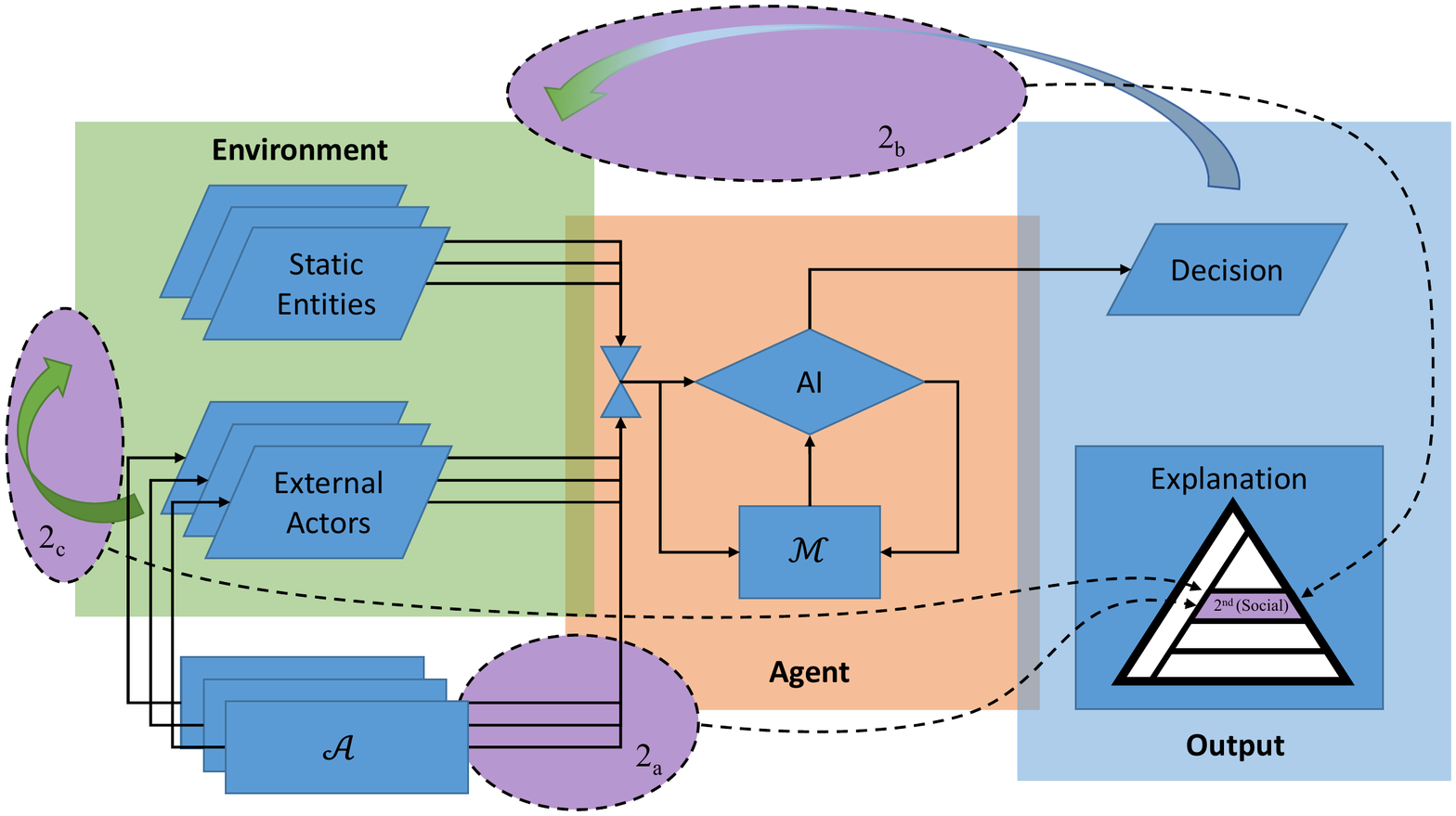}
  \caption{\textbf{Second-order (Social) Explanation} shown diagrammatically, where the purple ovals, $2_b$ and $2_c$ show that the agent's and other actors' behaviour affects the environment, which in turn affects the Merkwelt of the agent itself and the other Actors' Models ($\mathcal{A}$). The agent models the other actors' potential internal state and memory, represented by the oval $2_a$, to predict their future behaviour.}
  \label{fig:SecondOrder}
\end{figure}

This definition builds on Disposition explanation by also considering the agent's model of other actors' internal mental states. This idea is represented visually in Figure \ref{fig:SecondOrder}, where the three purple ovals indicate the areas of interest, $2_a$, $2_b$ and $2_c$. In order to provide this level of explanation, an agent must first assume that other actors in the environment have their own Merkwelt\footnote{Previously, we considered the agent's Merkwelt separately from the external perceptual influences on its Merkwelt because AI-systems are typically capable of separating these influences on their behaviour. In contrast, when constructing a predictive model of external actors it is assumed that such a separation of influences on those actors can not be readily determined. Therefore, it is assumed that the entirety of each external actor must be modelled.}, $2_a$, that guides their behaviour. The agent must also be engineered, or able to derive/learn through observation, how changes to the environment by the agent, $2_b$, and other actors, $2_c$, may influences each actor's internal state. It is this predictive model, which in turn feeds the agent's decision-making, that must be explained by a Social explanation. This model of how each separate actor is expected to behave is essentially a model of that external actor's Merkwelt, or at least how that manifests in its outward behaviour, and will be referred to in this paper as the \textit{Actor's Model} ($\mathcal{A}$). It should also be noted that this can also include a model interpreting the agent's own mental state as it believes others would see it.

This level is referred to as Social explanation as it aligns with basic human reasoning in social interactions, whereas the lower levels of reasoning simply involved an agent responding to its environment and achieving defined tasks. People's only interaction with such systems is to maintain the assigned tasks and/or take care to avoid the agent's operating environment. These systems work well in domains such as: autonomous packing facilities, where there are no people involved for the agent to interact; or, domestic vacuum cleaners, where people can reasonably avoid the devices' wanderings. In these contexts of low-level reasoning, Reactive and Disposition explanations will suffice. However, higher order reasoning is required for devices such as robot carers, autonomous self-driving cars, autonomous personal organisers, etc. Agents performing these higher levels of decision-making must be able to explain the reasoning involved in making those decisions. For a user to understand that reasoning, any explanation should also provide the AI's reasoning about other actors' dispositions. 

Social explanations allows an agent to answer questions like `why did you slow down when the pedestrian approached?'; open questions that do not imply a level, such as `why did you move my meeting time?; or, counterfactuals like `how would you have played if I had raised by a larger amount?'. The answers provided by such a system would go beyond its own intentions and explain how other actors' affected its behaviour. For instance, for the last three questions, it may provide explanations like: `I slowed down because the pedestrian approaching the intersection appeared intent on crossing'; `I changed your meeting time because I believed you would not wish to get up early after watching the late football game the night before'; or, `Had you raised by more then I would have thought you were bluffing'.

There are a number of factors in developing systems that can cognitively build a model of another actor. For instance, this may involve: the attribution of disposition or situational contexts to an actor's behaviour; cultural modelling of an actor; reflection on one's own beliefs and biases; emotion modelling of actors including the agent itself; and, the causal attribution of an actor's actions, or lack there of, to events or non-events. Furthermore, this is highly dependent on the type actor being modelled, for instance a person compared to an animal or another autonomous agent require different modelling approaches. Each situation involves developing a model for each actor's internal state or memory, including the agent's own as others may believe it to be. This modeling of external actors is an active research field with substantial work already caried out. For instance, culture modelling \citep{CultureModelling_Mascarenhas2016,CultureModelling_Hofstede2017}, user modelling \citep{UserModelling_Cawsey1993,UserModelling_Webb2001,PlayerModelling_Bakkes2012}, emotion recognition \cite{pal2019survey, mehta2019recent, rajan2019facial, chatterjee2019human, marechal2019survey, noroozi2018survey, supriya2019survey, li2018deep, salah2019video}, action recognition \cite{herath2017going, chen2017survey, cheng2015advances, dawn2016comprehensive, zhang2019comprehensive, singh2019human, kong2018human}, pedestrian prediction \cite{al2007modeling, gandhi2007pedestrian, gandhi2006pedestrian, hirakawa2018survey, rudenko2019human, wang2019pedestrian} and human intention modelling \cite{nocentini2019survey, ravichandar2016human, truong2019social, dutta2020human} are just a few areas that could be considered as modelling external actors.


There are a number of examples of applying IML based techniques to these Actor Models. One common application is to use XAI approaches to interpret a video stream to explain what had occurred or to provide captions \cite{mogadala2019trends, aafaq2018video, aineto2019model, xu2019joint, roy2019explainable}, or a narrative \cite{li2019emotion}. Alternatively, researchers use Local Interpretable Model-Agnostic Explanations (LIME) \cite{ribeiro2016model} to generate explanations of a range of Actors' Models, such as \cite{mathews2019explainable, weitz2019you}. As mentioned, all of these XAI systems are using IML techniques to interpret the conclusion made by the model. There are also a number of attempts to integrate Actors' Models to help direct agent decision-making. For instance, \cite{hao2019emotion} detects a person's current emotion, which is then used to guide the agent's behaviour --- the aim being to maintain a positive emotion in the human. This is not just limited to measuring external Actors' Models, but also the agent's own, for instance, approaches that determine a network or agent's own internal `emotion' or `hormones' that are in turn used to guide learning \cite{khashman2008modified, yang2012hybrid, thenius2013emann, yu2019emotion}. This process is often referred to as emotion augmentation and is discussed across multiple surveys \cite{balkenius2000computational, pentland2005socially, stromfelt2017emotion, moerland2018emotion, schuller2018age}.

However, Social explanation is not particularly interested in these methods \textit{per se}, but rather, in providing an explanation for how these methods' conclusions contributed to the agent's behaviour. A system providing an explanation of how it interpreted an external Actor's Model is essentially still only providing a Reactive explanation of that predictive model. This is because these approaches are applied domains for general machine learning and artificial intelligent algorithms. An XAI system providing a Social explanation may include the interpretation of the actor's model, but more importantly will describe how that outcome affected its final decision. Viewing Social explanation in this way is aimed at simplifying how such a system can be.

Despite the prevalence of actor modelling and augmentation there is relatively little work taking the next step of providing an explanation of how the augmented model has affected the agent's behaviour. In our discussion on Disposition explanation we discussed the recent introduction of Emotion-aware XAI (EXAI) \cite{kaptein2017role}. This paper not only discussed explaining the agent's internal emotions but also introduced the idea of explaining external actor's emotions and how they affected the agent's decision-making. More broadly than emotion augmentation, there has been work in mental state abduction \cite{sindlar2008mental, sindlar2009explaining, sindlar2011programming} of BDI agents, where there are examples of explanations describing how these deduced mental states have changed an agent's behaviour. BDI research in this area is primarily focused on identifying the mental state of another BDI agent, for which it already knows the possible mental states the agent can have, rather than modelling the mental state of less deterministic external agents, such as humans.


\subsection{N\textsuperscript{th}-order (Cultural) Explanations}
\label{subsection_nthorder} 

The social level, section \ref{subsection_second}, in human terms represents a relatively simple degree of interaction. People automatically expect others to determine what they are doing and adjust their behaviour accordingly. The issue is that this level of reasoning makes no assumptions of the need to provide some degree of \textit{quid pro quo}. That is, if one person gets out of another's way, is there a reciprocal responsibility? In a strongly hierarchical society --- perhaps not. In many human-agent interactions some may assume that we want such hierarchies --- where agents always adjust behaviour to stay out of people's way. However, this could result in agents being unable to complete required tasks as they must always avoid people. As most agent tasks are being carried out as proxies for a person, that person may not wish for their agent to always be subservient and defer to humans. In our motivating example, imagine that the autonomous car always gives way to all other cars in every situation. Such a car may never go anywhere and would quickly become useless to the human passengers. Furthermore, it will not behave like human drivers and, thereby, create confusion and risk as it does not meet other drivers' behavioural expectations.

In a complex society people do not just develop a model of other's and then adjust their behaviour to avoid them. Instead, they also derive a set of \textit{expectations} of how those other actors' believe they will behave to avoid them. For instance, if two people walking through a shopping center are heading directly towards each other, person A may decide that they should side step to the left slightly while also having an expectation that person B will do similarly creating enough distance between their paths to avoid the impending collision. This level of reasoning is equated to third-order intentionality \cite{dennett1983intentional} because person A, not only has a model of what person B will do, but recognises that person B will expect person A to do likewise --- and vice versa. This represents an ever increasing recursive level of mentalisation \cite{holmes2008mentalisation, lewis2017higher} indicating an understanding of a set of cultural rules about behaviour. For instance, the above example was based on the cultural assumptions used in Australia --- if person A is visiting the United States of America, where people step to the right instead, taking a step to the left is likely to create confusion and possibly be the cause of a collision. This represents a shared cultural and social awareness. With people that know each other personally this same awareness may go beyond general cultural norms and consider a deeper and personal understanding of expectations. 

Current First-order intentional agents like vacuum robots are built with the assumption that people will entirely defer to them and get out of their way. People tolerate this due to our low level of expectation of the device's intelligence and because avoiding one such device is simple. However, with more devices involved in more complex interactions greater expectations will be placed on the agent to not only avoid us, but to culturally integrate with us. Therefore, agents will be required to, not only be provided with or learn a model of how other actors will behave, but to also determine what expectations those actors may have of how the agent should behave. Ultimately, this is recursive, with an agent also needing to model, not just other actors' behaviour, but also their model of how the agent itself may behave (Fourth-order intentionality). Any process used to reason about other actor's expectation of the agent's behaviour, and thereby altering its behaviour accordingly, should also be explainable in conjunction with lower-order explanations. Collectively we will formally categorize these as N\textsuperscript{th}-order explanations, but will generally refer to them by the more descriptive name of Cultural explanations:

\bigskip
\begin{addmargin}[4em]{3em}
\textbf{N\textsuperscript{th}-order (Cultural) Explanation:} is an explanation of a decision made by the agent based on what it has determined is expected of it culturally, separate from its primary objective, by other actors.
\end{addmargin}
\bigskip

\begin{figure}
  \centering
  \includegraphics[trim={2.9cm 4.1cm 2.2cm 3.6cm}, clip, width=\textwidth] {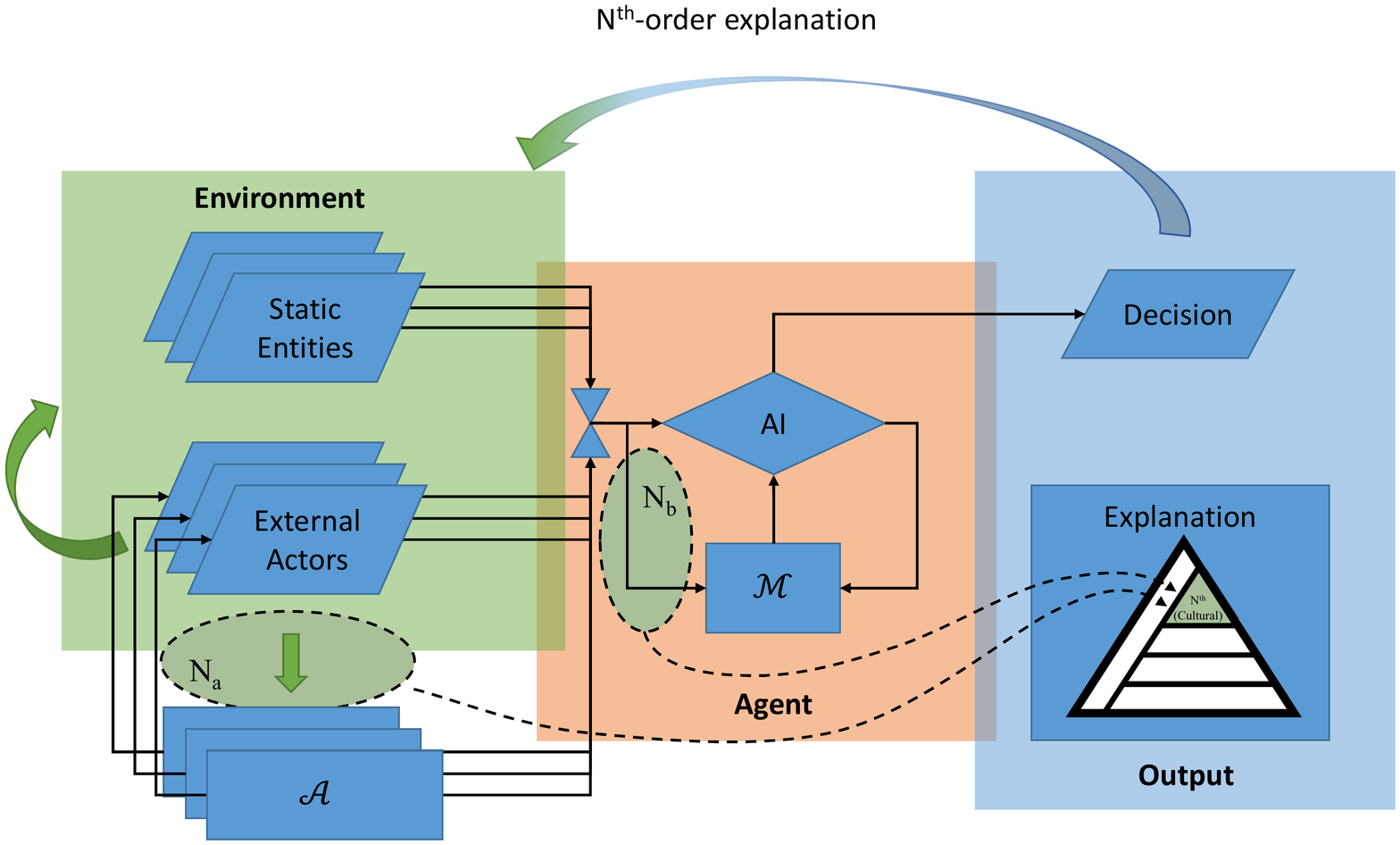}
  \caption{\textbf{N\textsuperscript{th}-order (Cultural) Explanation } shown diagrammatically, where the green oval, $N_a$ indicates that we need to model how the actors' potential internal state and memory are affected by changes in the environment, while, $N_b$ indicates the need to explain how the changing environment affects the agent's Merkwelt.}
  \label{fig:NthOrder}
\end{figure}

This definition builds on Social explanations by also considering the agent's model of other actors' internal model of the agent's behaviour, which has been built or learnt through observation. In other words, a Cultural explanation details how the agent's behaviour is modified due to what it understand is expected of it by other agents. This idea is represented visually in Figure \ref{fig:NthOrder}, where the two green ovals, $N_a$ and $N_b$ indicate the areas of interest. Here it can be seen that we are interested in modelling both how the external actors' models are altered, $N_a$, as well as how the agent's own Merkwelt, $N_b$, is affected by changes in the environment. Such an explanation requires the agent, to not only model the other actors' potential internal state and memory, but also how the actor's model may be affected by changes in the environment.

This level is also referred to as Cultural explanation as it aligns with basic human reasoning in cultural interactions and is aligned to the philosophical theory of Sociality: incorporating the idea of a collective intentionality \cite{tuomela2007philosophy}. This implies that any expectations determined for one Actor's Model or situation may potentially be transferable or shareable with other Actor's Models or situations. The types of questions answered by this level are the same as those in Social explanations, except now the provided explanation incorporates its cultural understanding of others' expected behaviour. For instance, a cultural understanding of one road user in France can, for the most part, be transferable to all road users in France, but only partially transferred to road users in China. Cultural explanations would, therefore, be able to provide explanations such as: `I waited a moment to hold the door open for the person behind me, as social obligations indicate that they would expect this behaviour'; or, counterfactuals like `I didn't slow down for the pedestrian because road laws state I have right of way, and therefore, I assumed they expected me to continue'. These explanations can also be compounded by including explanations of multiple actors' expectations. For instance, the last example could be extended to be `I didn't slow down for the pedestrian because: road laws state I have right of way, and therefore, I assumed the pedestrian would expect me to continue; and, the car behind would expect me to continue at my current speed'. 

As previously discussed in section \ref{subsection_second}, an agent that is modelling another actor is ultimately attempting to predict that actor's behaviour. In contrast, an agent that is designed to respond to cultural expectations must predict what the actor expects the agent to do and how they are likely to respond to the agent's behaviour. This form of behaviour modelling has been widely studied. For instance, Normative Agents research has extensively studied multi-agent fields like BDI agents \cite{adam2016bdi, santos2017detection, hollander2011current, beheshti2014normative}. These approaches focus on constructing multi-agent societies where social norms (of the multi-agent society) can be applied to the expectations of the agent. However, like other BDI work these systems typically avoid operating in dynamic human environments. Game Theory \cite{myerson2013game}, which models the strategic interaction between rational decision-makers, assumes that other actors will behave optimally to achieve their goal. Game Theory has been extensively studied and applied to numerous real-world domains including modeling social and cultural interactions \cite{camerer2011behavioral, suleiman2012tools}. One common use is in modelling opponents in games. For example, Alpha Go \cite{silver2016alphago}, which defeated the World Champion Lee Sedol in 2016, utilised a Monte Carlo tree search (MCTS) \citep{silver2016mastering} to predict responses by its opposition on various possible actions it could make. Furthermore, it can build a probability to do this over a sequence of actions and not simply model the next behaviour. Modelling expectations has been more directly modelled in social action research, which models \textit{when, why and how} external demands on an agent affect its goals or actions \cite{castelfranchi1998modelling, conte2016cognitive, poggi2010cognitive}. Other learning systems, such as Reinforcement Learning are sometimes designed to incorporate social and cultural awareness of the environment into their action selection, for instance, when navigating through crowded domains \cite{charalampous2017recent, chen2017socially, triebel2016spencer, kim2016socially, vasquez2014inverse, ritschel2018socially}.

Nevertheless, these areas have not currently attracted significant research interest in the development of explanation. For instance, David Silver, DeepMind Researcher, reportedly had no insight into why AlphaGo made such a creative play as move 37 in Game 2, which surprised both commentators and Lee Sedol, until he had investigated the actual calculations made by the program \citep{Metz2016deepmind}. For these systems to go the next step and be used by everyday \textit{non-expert} users, that are not able to inspect an agent's value-function, agents must be able to provide explanations at this level. Recent research by \citeauthor{kampik2019explaining} \cite{kampik2019explaining} into explaining sympathetic actions that incorporate utility for socially beneficial behaviour at the detriment of personal gain and using this to explain behaviour is an exciting example of culturally-aware explanation. Other examples of explanation systems based on cultural expectations do not align to a unique field of research, but are occasionally published under more general terms such as understandability \cite{hellstrom2018understandable}, transparency \cite{wortham2017robot}, predictability \cite{dragan2013legibility}.

Interestingly, when an agent begins to base its decisions on this idea of \textit{quid pro quo}, it not only means there are expectations on the agent but also on the other actors. Like all socially-aware actors they will also have expectations to follow. Understanding and modeling this means an agent can go beyond simply predicting how an actor's model will change based on the agent's behaviour, and instead, model how it can actively change the actor's behaviour. Research in Computers As Persuasive Technologies (or Captology) investigates approaches to persuade external actors. Captology is already used in a number of application domains, such as commerce, safety, environment, productivity, disease management, health care, activism and personal management \cite{fogg2007motivating, albert2004health, nemery2011use, conway2019improving, rist2019promoting}. Captology implies a goal or intention to alter and change other actors' internal beliefs, motivations and behaviours and can be considered as Fourth-order intentionality \cite{dennett1983intentional}. An explanation for such a system would amount to a description of how an agent's action was based on the goal/intention to change an actor's beliefs, motivations and/or behaviours in a particular way. In this way the explanation could be generated using a Disposition explanation approach. We collectively have wrapped these terms up into Cultural explanation in this paper as levels of intentionality, in theory recursively continue. To our knowledge there have been no attempts to provide explanations for such systems.


\subsection{Meta (Reflective) Explanations}
\label{subsection_meta} 

One of the major drivers in the development of XAI is to provide understanding to people so they can trust the decisions made. However, a major issue in XAI research, which is generally not discussed, is if we cannot trust the agent's original decision, how can we trust the agent's explanation of that decision? The majority of research in XAI has been researcher/developer-centric \cite{Ehsan2019OnDA} --- providing an explanation that focuses on interpreting the decision. As these are direct interpretation methods they tend to be more readily acceptable to an expert, but do not generally provide the level of understandability required by a non-expert. Human-centered interaction methods instead approach the problem from the human perspective by focusing on the user's acceptance and understanding \cite{abdul2018trends, Ehsan2019OnDA, ehsan2019automated}. These approaches can tend to focus on improving the understandability of the explanation, however, this may be at the detriment of accurately reflecting the implementation details behind the decision. A human-centered approach to XAI increases the potential for the provision of untrustworthy explanations. In this section we will briefly discuss the three primary issues we see in the provision of understandable explanations:

\begin{enumerate}
   \item Utility-driven XAI --- using human acceptance and understanding to assess the success of an agent's explanation.
   \item Deceptive explanations - where an agent needs to hide its true objective from the explainee. 
   \label{issue_2}
   \item Simplified/generalised explanations - inadvertent deception through omission. \label{issue_3}
\end{enumerate}

Discussed in section \ref{subsection_Conversation} the \textit{narrating-self}, the conscious part of the mind, creates explanations that support the internal goals of the individual. Therefore, the aim of the explanation could be considered in terms of maximising the agent's utility. This aligns with a human-centered approach to explanation, where the aim is to measure the \textit{utility} of an explanation on the human's perceived level of acceptance or understanding and to use this to refine the explanations provided \cite{Ehsan2019OnDA, ehsan2019automated, mclaughlin1988utility, kim1991explanation}. In fact, the primary objective when providing an explanation could be considered the requirement to satisfy the explainee's need to understand and accept a decision --- not to accurately reflect the actual decision-making process. This requirement to provide an understanding to the user is potentially in direct conflict with providing a `truthful' and accurate explanation. An agent that learns how to explain a concept may learn to use an explanation that returns the highest utility, rather than the most accurate. For example, Emotion-aware XAI was introduced as an approach to both explain an agent's internal disposition (section \ref{subsection_first}) as well as, interpreting an external actor's disposition (section \ref{subsection_second}). One approach used to develop such agents is to use emotion-driven or emotion augmentation learning methods \cite{marinier2008emotion, elliott1998model, marinier2009computational, hoey2016affect, gadanho2001robot, hao2019emotion, yu2019emotion}. These methods often use the emotion of the agent or the perceived emotion of an actor to control its behaviour, such as an agent's reward being aligned to pleasing its user. If this approach is used when providing explanations, then an agent will learn an explanation that satisfies the user --- not the truth.




\citeauthor{person2019agents} \cite{person2019agents} recently introduced the idea of rebellious and deceptive explanations. These are explanations that are generated, not just to explain a decision, but to intentionally persuade a person and, thereby, change their internal beliefs and desires. Such explanations are not frequently discussed in the literature, but represent a significant safety risk to one of the primary objectives of XAI --- developing trust. Deception can be in the obvious form of lying \cite{van2014dynamics, sakama2011logical, van2012logic}, but as \citeauthor{sakama2015formal} \cite{sakama2015formal} suggests it can also be a result of bluffing or truth-telling (agent makes a truthful explanation that deceives the listener) \cite{sakama2015formal, sakama2010many}. Intentionally deceiving agents for instance can be used during negotiations, such as where an agent representing a company may need to deceive a customer to advantage the company. Current approaches \cite{nguyen2011asp, zlotkin1991incomplete, sakama2011logical} focus on logic formalisms for deception, but future work in explaining black-box Machine Learning algorithms are likely to also encounter this concept in future research. \citeauthor{sakama2014formal} \cite{sakama2014formal} suggest four \textit{Quantitative and Qualitative Maxims for Dishonesty} that build on from \citeauthor{grice1975logic}'s \cite{grice1975logic} \textit{maxims} for common and accepted rules of conversation. \citeauthor{sakama2014formal}'s \cite{sakama2014formal, person2019agents} \textit{maxims} suggest an agent should:
\begin{enumerate}
   \item Lie, Bullshit (BS), or Withhold Information (WI) as little as possible to achieve its objective
   \item Never lie if it can achieve its objective by BS.
   \item Never lie, nor BS if it can achieve its objective by WI.
   \item Never lie, BS, nor WI if it can achieve its objective with a half truth.
\end{enumerate}

Deception does not necessarily need to be designed - it could also be a learnt trait of an agent. The last paragraph of section \ref{subsection_nthorder} discussed the research field of Captology, which can be considered a Fourth-order intentional systems. These systems are designed to persuade and alter an actor's beliefs, motivations and/or behaviours. This field of research was introduced as an example of the recursively higher-levels that a Cultural explanation system ultimately may need to explain. However, it also introduces a troubling concept: what if the agent's method of persuasion is through the explanation itself? 


A deceptive explanation does not only result from an agent having the intention to deceive, or to maximise a utility for human acceptance and understanding, but can also be the result of a poorly formed explanation. For instance, Grice's \cite{grice1975logic} \textit{maxims} for common and accepted rules of conversation, suggests an explanation should be as brief and as general as possible. One cause of miscommunication, and potentially unintentionally deceptive explanations, is through the omission of some information to maintain brevity, causing a misunderstanding on the receiver's end. The process of generalisation can also result in the explanation misrepresenting certain facts about a situation. 


In this paper we suggest that human trust requires more than just the explanation, but also requires an understanding of the process used in formulating the explanation. That is, how was the explanation generated or selected, as opposed to other explanations, and importantly, how the explanation provided is related to the decision made. We suggest that a truly broad-XAI agent would need to be able to perform a reflective exercise on its own explanation process. Formally this is categorized as \textit{Meta-explanation} and descriptively referred to as \textit{Reflective explanation}:

\bigskip
\begin{addmargin}[4em]{3em}
\textbf{Meta (Reflective) Explanation:} is an explanation detailing the process and factors that were used to generate, infer or select an explanation.
\end{addmargin}
\bigskip

\begin{figure}
  \centering
  \includegraphics[trim={2.9cm 4.1cm 2.2cm 3.6cm}, clip, width=\textwidth]{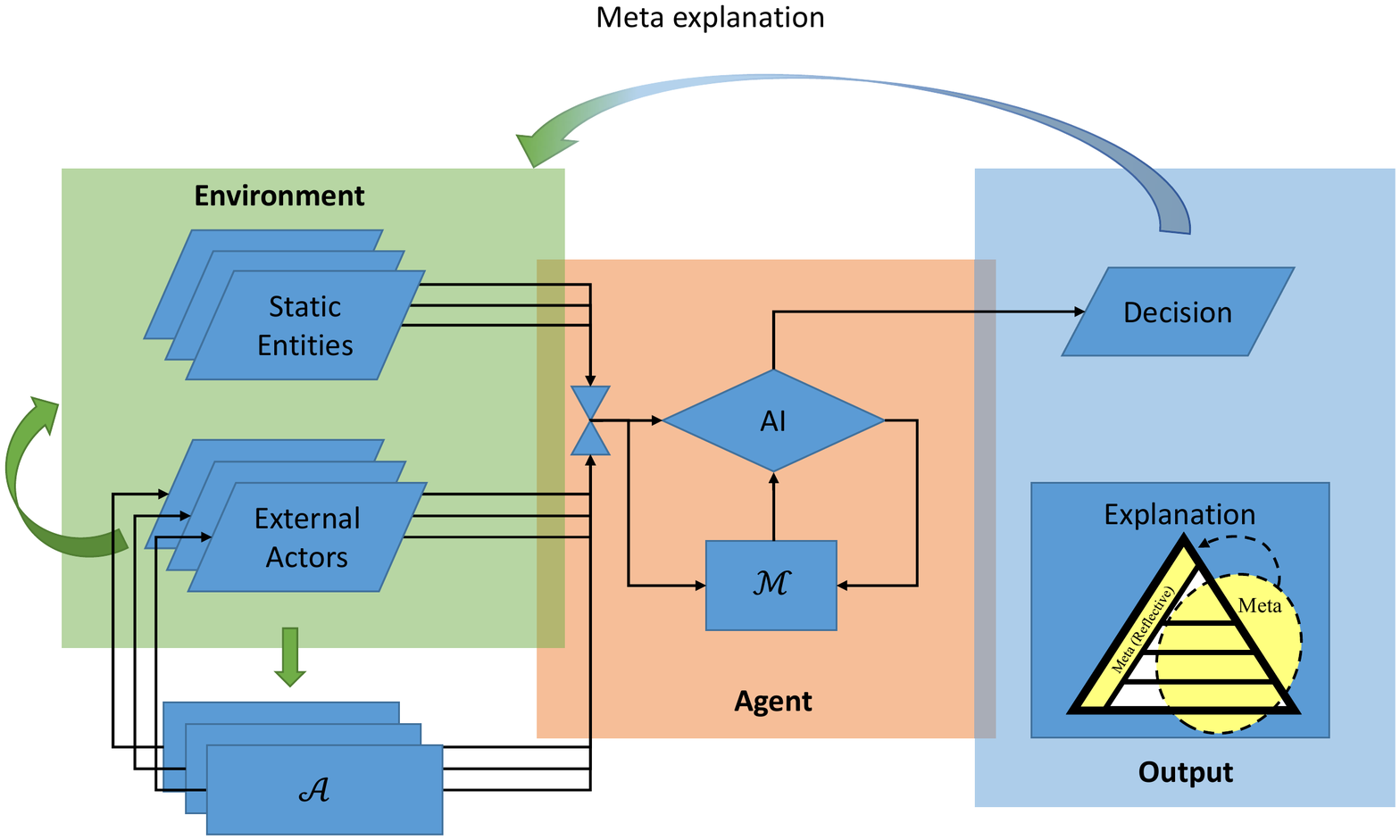}
  \caption{\textbf{Meta (Reflective) Explanation} shown diagrammatically, where the yellow circle, \textit{Meta}, encompasses the explanation levels indicating that a Meta-explanation is concerned with inferring how each of the explanation levels were mapped.}
  \label{fig:meta}
\end{figure}

Meta-explanation is not a commonly used term in the XAI literature. Generally, it has been used to refer to an explanation of Meta-knowledge \cite{pitrat2006meta, galitsky2016formalizing}. That is, rather than explaining why a decision is the correct decision (the explanation) it explains the problem solving process in solving it. In systems based on logical inferences this often referred to an explanation based on a rule-trace \cite{galitsky2016formalizing}. More broadly than just AI, the literature has also separated explanation into two levels: Object-level explanation and Meta-explanation. The Object-level provides an explanation of particular decisions or arguments, whereas a Meta-explanation references the scenario structure behind the explanation or argument and may include historical decisions or justifications \cite{Galitsky2010ExplanationVM}. Up until now, we have suggested four levels of explanation that equate to the Object-level explanation used in the wider literature. A Meta-explanation equates to the wider literature use of the term except that we propose to broaden these ideas specifically for use in XAI and suggest that a Meta-explanation, not only describes the problem solving process or the scenario structure, but also the process used to select or generate the explanation of the original decision. This is illustrated diagrammatically in Figure \ref{fig:meta}, where all explanations previously discussed represent the internal processes requiring explanation. Therefore, a Meta-explanation, the yellow circle, \textit{Meta}, is drawn from all other forms of explanation. 

Meta-explanation was deliberately illustrated in Figure \ref{fig:Levels} as running across and alongside each of the explanation levels. The aim was to indicate that a Meta-explanation can form a symbiotic relationship with each of the levels. Essentially, we are suggesting that an explanation may be just a level explanation or could also contain part or entirely Meta-explanation details. For instance, a Reactive explanation of an agent with Zero-intentionality is also considered a Meta-explanation. This is because a Reactive explanation is a direct interpretation of the algorithms processing of its reaction to an input. The objective of the explanation is to illustrate the problem solving process. This is a traditional form of Meta-explanation that is often called Interpretable Machine Learning (IML) \cite{guidotti2018survey, doshi2017towards}. For example, asking an agent why it claimed to have seen a hat in an image would require the system to identify where in the image it saw the hat or the process used to conclude the classification of a hat through a neural-trace, rule-trace, decision tree path, etc. This is a Reactive explanation presented using a Meta-explanation. 

A Reactive explanation that is embedded within a Broad-XAI explanation (an explanation spanning multiple levels) is more likely to indicate what it saw, but not how it reached that conclusion. In such an explanation, the question of how it generated the classification is less important than the interplay of how that affected other levels of the explanation. This is due to Grice's \cite{grice1975logic} suggestion to simplify an explanation and, thus, representing a Reactive explanation without including the Meta-explanation. For example, in the self-driving car example, in section \ref{subsection_example}, the first explanation suggests that the agent did not see a person pushing a bicycle. This is a Reactive explanation without a Meta-explanation. Should the explainee require further information about how the car reached the conclusion that there was no person with a bicycle they would need to request a deeper explanation such as a Meta-explanation of the process used by the agent. 

This separation of level-explanation and Meta-explanation is clearer in the higher-levels as these are typically already providing Broad-XAI explanations, and therefore, asking how it determined an explanation becomes more meaningful. For a fully Broad-XAI explanation that goes across all levels of the hierarchy providing a Meta-explanation becomes an important component of drilling down to validate the process used by the agent in determining the explanation given. For instance, if the self-driving car gave the third example explanation that it thought the person was giving way then the Meta-explanation would need to provide an explanation of the agent's process used to determine the person was giving way. 

The exact nature of a Meta-explanation is directly determined by the method used to generate the original explanation. For instance, if an explanation is determined from a set of predetermined options then the Meta-explanation will provide details of the classification process used. Then again, if the explanation uses probabilities based on past situations then the Meta-explanation would need to detail or trace the approaches used to determine the probabilities calculated. Essentially, a Meta-explanation could be considered as the \textit{backing} to the \textit{warrant} (explanation) of Toulman's argumentation model \cite{toulmin_2003}. 

Given this definition of a Meta-explanation as being purely an interpretation of the process used in forming the explanation then the Meta-explanation is also disconnected from the agent's learning, memory and reward structures. This means the resulting explanation can be seen as a truthful (accurate) reflection of the agent's behaviour. This is an important differentiation as it separates a Meta-explanation from the first issue raised by utility-driven XAI. A regular explanation is aiming to satisfy the explainee's need to understand the reasons for the decision and not the processing method used. If meeting this need simplifies, generalises or omits information about the agent's true objectives then the additional provision of a Meta-explanation, while potentially less understandable, provides an honest and accurate reflection of the decisions and explanations provided (Issue 3). To this end we actually see such a Meta-explanation as outside the normal conversational component of an explanation and as a method of providing a deeper understanding to the explainee if required. 

Unfortunately, we believe it is unlikely that a Meta-explanation will be provided by certain systems. Intentionally deceptive agents (issue \ref{issue_2}) created by organisations will frequently require the exact decision-making process and its associated explanation generation to be hidden from clients. Forcing companies to provide a Meta-explanation is an issue to be taken up with governments --- potentially the same ones that will be unwilling to provide such details themselves. For instance, the new General Data Protection Regulation \cite{goodman2016european} aims to ensure such processing components are made available to those affected by decisions made by automated systems. There is a risk that companies could view the provision of a regular explanation as sufficient. Instead, we believe that a Meta-explanation of these decisions are also required.

\section{A Conceptual Model for Providing Explanations Through Conversation}
\label{section_ConceptualModel}

The aim of this paper has been to present a conceptual approach to bringing together different AI and XAI components to guide and inspire future research towards the development of \textit{Broad-XAI} based conversational explanations. In so doing we have reviewed a number of disparate research fields and through a structured set of \textit{levels of explanation} illustrated how these separate fields can provide a unified approach to XAI. In particular, this paper has rejected current interpretable machine learning (IML), or debugging-based, approaches as a solution in-and-of-themselves for XAI. We have argued that these approaches are primarily focused on providing Zero-order (Reactive) explanations. Higher levels are identified that allow for more socially and culturally-aware explanations to provide a human-aligned explanation.

This paper has focused on providing a more philosophical-based view of \textit{Broad-XAI} and implementing such a system is well beyond the scope of this paper. However, in this section we will outline a conceptual model of how such an implementation could be carried out using the suggested \textit{levels of explanation} discussed in this paper. This will be described in a general fashion, such that specific implementations could be developed, tailored to particular application requirements. In so doing we will brush over the minutia of such an implementation, as these should be the focus of future research. The following sub-section will reprise the earlier discussion on explanation as conversation and present a process for performing an explanation through an interactive conversational process. Subsequently, the following sub-section will provide an insight into the AI-model required to provide the cognitive processes required to for Broad-XAI and the identified levels of explanation. The final sub-section will provide a brief review of current approaches to developing Broad-XAI approaches and discuss how they relate to this paper's proposed approach.



\subsection{A Broad-XAI Conversational Process}
\label{subsection_SocialProcess}

In section \ref{subsection_Conversation}, we discussed how an explanation was a process consisting of both a \textit{cognitive} and \textit{social process}, where the social process involves an interaction between the explainer and the explainee. The aim of this interaction is to convey the explainer's proposed causes of an event or outcome such that the explainee understands and accepts the explanation. This interaction forms the fundamental human process of communication, which in turn informs the explainee's cognitive process. On the explainee's side of this communication the identified causes and counterfactuals are aligned to the explainee's currently accepted understanding of the world. The explainee can then accept or reject the provided explanation. Alternatively, they may require the agent to provide backing claims \cite{antaki1992explaining} by requesting additional, more specialised or detailed explanation to resolve any identified internal conflicts. 

Furthermore, it was identified that the psychological models of explanation in humans uses a reductionist model (see Figure \ref{fig:psychologyLevels}) suggesting explanations are provided and received using a top-down approach. In contrast, we have argued that AI systems do not operate in the same socially-focused cognitive process and instead utilise a data-driven constructivist approach. We, therefore, presented a set of levels that builds explanations from the bottom-up. Ultimately, the explainer must still relate the explanations to the explainee's expectations. Therefore, it is important to conclude with a discussion of how the two models work together to communicate an explanation. 

\begin{figure}
  \centering
  \includegraphics[trim={5.2cm 4.4cm 5.2cm 4.4cm}, clip, width=34pc]
  {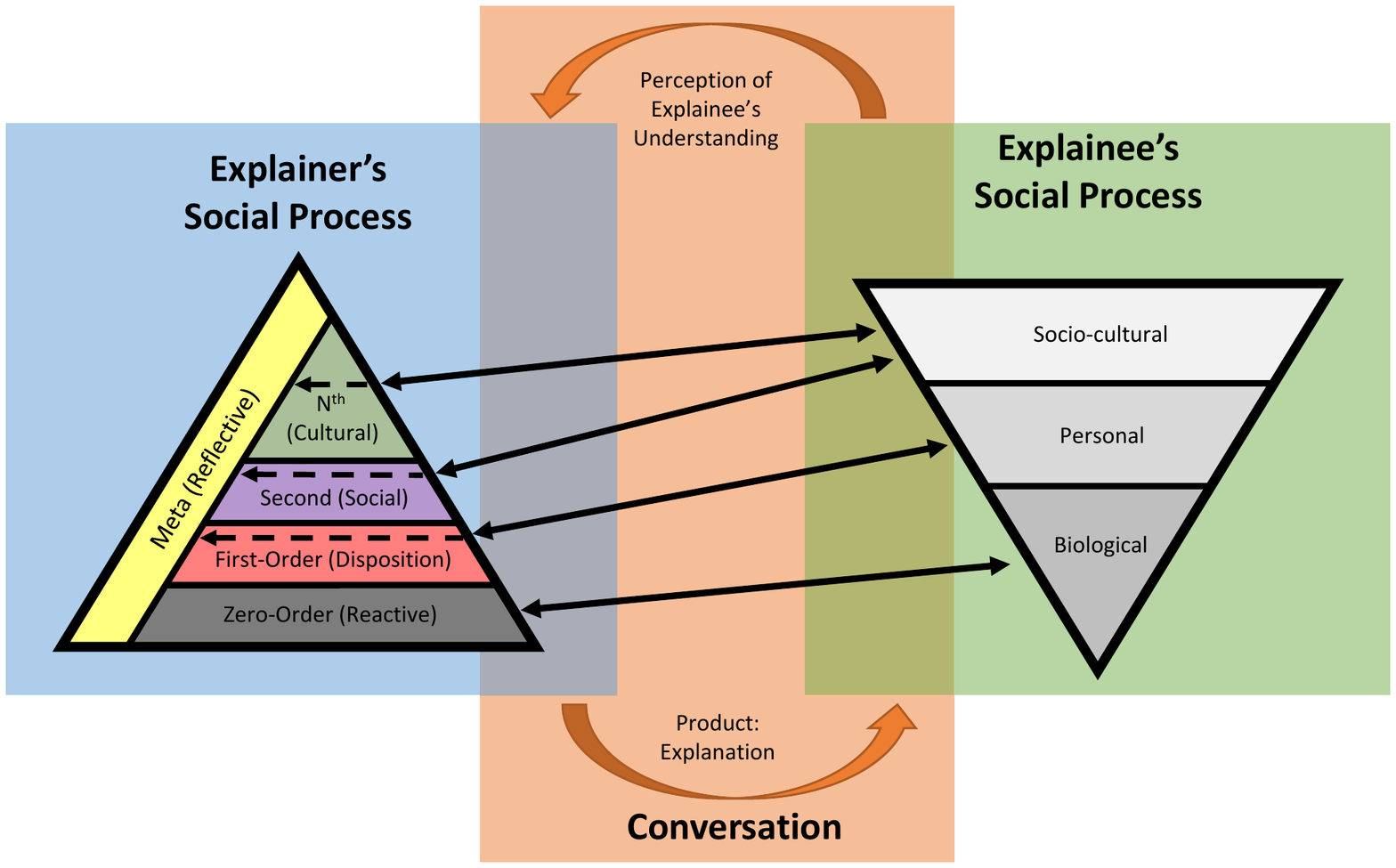}
  \caption{\textbf{Conversational Interaction} between the levels of Explainable AI and human psychological levels (builds on top of Figure \ref{fig:Conversation}). When the explainee requests an explanation (black arrows) at a Socio-cultural level the agent responds from a Second or N\textsuperscript{th}-order level of explanation. Similarly, when requesting an explanation at the Personal or Biological level the agent will respond from a First or Zero-order level of explanation respectively. In each case, the explainee can also request a Meta-explanation (grey arrows) on how a prior explanation was generated.}
  \label{fig:Reprise}
\end{figure}

In the motivating example of the autonomous self-driving car, we considered a number of plausible explanations. Each of these are drawn from the various levels of explanation given throughout the paper. One approach to provide an explanation is to identify the appropriate level from the approach used by the explainee when phrasing the request for an explanation. Figure \ref{fig:Reprise} provides an illustration of how an agent's levels of explanation aligns with that of a person's psychologically-based levels by embedding these models within the conversation process shown earlier in Figure \ref{fig:Conversation}. For instance, let us assume the user asks `what did the vehicle perceive just prior to the collision'. Clearly such a request is for a Reactive explanation of the agent's perception to determine the reason for the reaction. Generally speaking, we expect it would be unlikely that the user will phrase a request for a specific level. Rather, a more likely assumption would be that the explainee would ask `why didn't the car give way to the pedestrian pushing a bicycle'? Such a question is not sufficiently aligned to any individual level of explanation. In fact, any of the four responses indicated in section \ref{subsection_Conversation} would provide a reasonable explanation --- although alone they are unlikely to fully satisfy the explainee. 

One approach would be to wrap-up all the explanations from each level into a single explanation. For example, a combined explanation may look something like: 

\bigskip
\begin{addmargin}[4em]{3em}
`The system perceived the pedestrian with a bicycle approaching, however, was motivated to remain in the right lane due to an approaching exit, believing the person would give way to the vehicle and, as the car had the right of way, that the person would expect the car to continue'.
\end{addmargin}
\bigskip

However, this would break Lombrozo \cite{lombrozo2007simplicity} and Thagard's \cite{thagard1989explanatory} suggestions that an explanation should rely on as few causes (simple) as possible that cover more events (general) and maintain consistency with peoples' prior knowledge (coherent). The combined explanation makes no determination of the explainee's prior knowledge; is very specific to the exact situation; and, complicates the explanation with many causes that may be irrelevant. Therefore, such an explanation, we believe, would not be suitable. 

\begin{figure}
  \centering
  \includegraphics[trim={0.4cm 4.1cm 0.6cm 4.0cm}, clip, width=33pc]{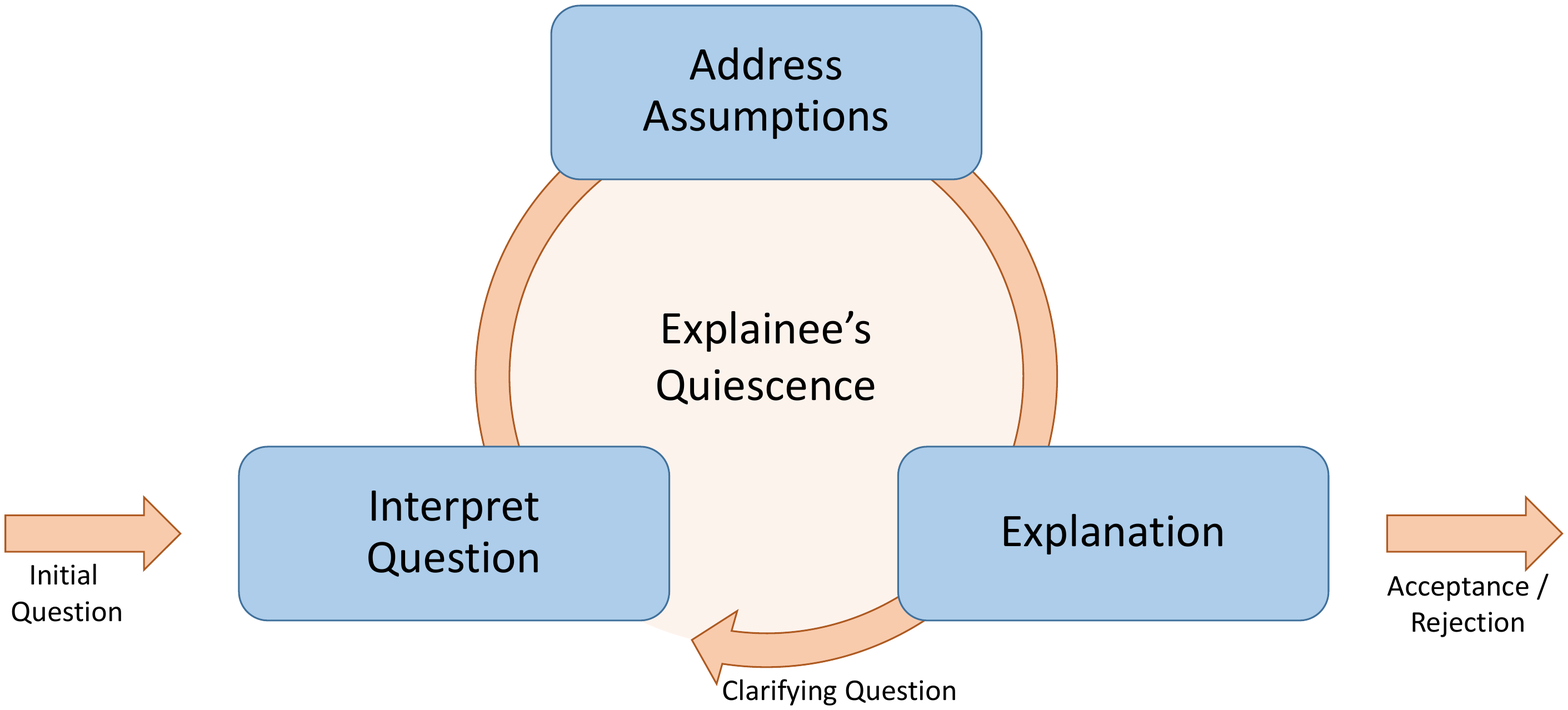}
  \caption{\textbf{The Conversational Process} iterates through three stages until the explainee reaches quiescence - A utility measure of understanding and acceptance. The cycle terminates when the explainee accepts or rejects the explanation and moves on to either a new line of questioning or no longer requires an explanation.}
  \label{fig:Quiescence}
\end{figure}

Interactionism in sociology studies the social process: between two people; between a person and an animal; increasingly, between a person and an AI-driven agent \cite{cerulo2011social, can2016human, elder2019living}; and, even between animals and an AI-driven agent \cite{coeckelbergh2011humans}. Interactionist theories suggest that multiple levels of explanation are necessary to explain a particular behaviour and that often the interplay between these levels is also required \cite{McLeod2019Reduction}. 
This suggests that an explanation should be constructed from several levels. We envision that this could be done in an incremental approach such as through an interactive conversation. An agent initially would provide an explanation at the highest level of the agent's intentionality that answers the question and drills down through the levels to answer subsequent queries until the explainee accepts that the explanation aligns with their prior knowledge. This aligns it with the psychological based model used by people, while preventing the provision of irrelevant causes. We see this as a three step recurring process for the agent, as illustrated in Figure \ref{fig:Quiescence}. This approach highlights that a conversational explanation is one that, at its core, is a utility-driven explanation, where the utility is measured by the amount of \textit{quiescence} in the explainee. Quiescence indicates the state of being quiet and in this paper is used to measure stability in understanding and acceptance --- a quiescent explainee no longer needs to ask any more questions as they have balanced the agent's reasoning with their own world view. However, the conversational explanation process can also terminate, even when quiescence is not reached, by the explainee rejecting an explanation entirely. Such a situation occurs when the explanation does not align with their personal world view. Similarly, the explainee may end the process because of frustration or to move on to a different line of questioning. The measurement of this type of utility has been considered through examination dialogues \cite{walton2006examination, arioua2015formalizing, walton2011dialogue}. This process equates to Fourth-order level of intentionality where the agent is actively trying to move the explainee's understanding towards a level of understanding and acceptance. 

The first stage is to correctly \textit{interpret the question} (either explicit or implicit) conveyed to the agent. On the surface this appears to be simply an interpretation of their words or body language. However, to provide a conversational explanation the agent needs to understand the underlying nature of the explainee asking the question. Therefore, we suggest the agent requires a model of the explainee's current understanding of the world in the form of an Actor's Model. Such a model allows the agent to correctly interpret the \textit{true} nature of their request for an explanation and to provide a more accurate explanation. This stage illustrates a relationship with Second-order intentionality, where the agent's choice of behaviour, the explanation, is determined by a model of the explainee. User modelling in Human Computer Interaction (HCI) \cite{biswas2012brief, biswas2010brief,nocentini2019survey,dutta2020human,truong2019social,ravichandar2016human} approaches can be used to model the explainee allowing more meaningful interpretations of the questions. This is evident in recent work in the emerging field of Personalised Explanations \cite{schneider2019personalized}. At the conclusion of this stage the agent should have a, possibly re-framed, question that is then addressed in the next stage.

The second stage aims to identify and confirm or clarify any \textit{assumptions} that exist in the question that may be incorrect. These are important because if the explainee has an assumption about the agent's reasoning that is incorrect then identifying that assumption and pointing out in what way it is wrong is likely to address their primary concern. For example, when asking `why didn't the car give way to the pedestrian pushing a bicycle?' there is an assumption that there was a pedestrian pushing a bicycle and that the car had detected this. If this was not detected then the agent should jump immediately to a Reactive explanation and identify what it did, or did not, perceive. Essentially this involves rephrasing the question to be specific to the lower level of intentionality ready for the third stage. If there are no assumptions in the question, or the assumptions are correct, then the question can be passed to the next stage.

The final stage is to produce an \textit{explanation} based on the highest relevant level to the question asked involved in the original reasoning process that has not yet been explained. Therefore, assuming the agent's original reasoning process involved all levels of intentionality and that the question is open to a response from any level then the agent will respond from the highest level of explanation. For instance, the car `believed that it had the right of way and that the person would expect the car to continue'. The elegance of this explanation is that, in a single generalised explanation, it has implicitly also confirmed that the assumption that there was a pedestrian pushing a bicycle was in fact perceived by the agent and the car had modelled both its responsibility and determined a prediction of the pedestrians expected intentionality.

Once this explanation has been offered the agent measures the explainee's quiescence. In the event that the explainee has not reached this state yet then we repeat the cycle with a new question. We suggest there are three forms for this subsequent question:
\begin{itemize}
  \item They ask a question that drills down further. For instance, `did the pedestrian show any sign they intended to give way?'. Such a question will generally require a lower level of explanation on the next cycle.
  \item Their body language or facial expression indicates some level of confusion resulting from the explanation given. In which case the agent can offer an explanation from a lower level to provide greater clarity.
  \item They ask an alternative branch of questioning. For example, `Why were you in the right lane?' A totally new branch indicates they have exhausted the prior line of inquiry and now wish to start a new explanation cycle. 
\end{itemize}

In this way, the agent can provide general explanations initially and progress towards more specific explanations until the explainee is satisfied. This approach addresses Miller's \cite{miller2017explanationBook, miller2017explainable} three key areas of interest in a quality explanation system: contrastive explanation, attribution theory and explanation selection. For instance, this approach builds a structural model of causal facts and uses these to provide contrastive explanations; while also providing detailed attribution of both its own, and its perception of other actors, feelings, beliefs and intentions; and, selects the appropriate explanation for the context required by the explainee. This process provides the impression to the explainee that the system is listening and responding in the context of their needs.

\subsection{An Underlying AI-based Cognitive Process}
\label{subsection_CognitiveProcess}

\begin{figure}
  \centering
  \includegraphics[trim={1.3cm 2.1cm 1.9cm 2.1cm}, clip, width=39pc]{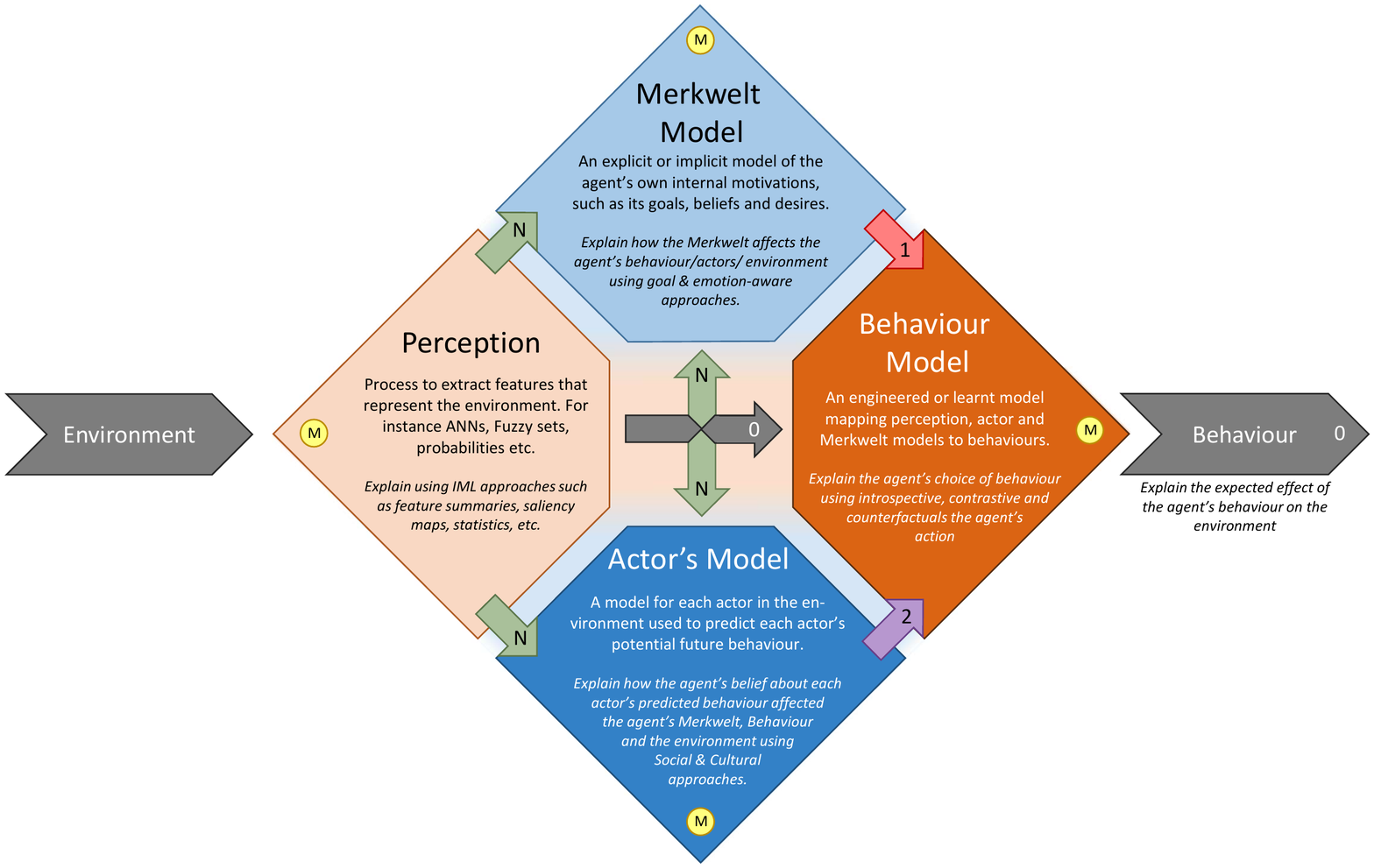} 
  \caption{\textbf{The Cognitive Process} representing the minimum components required for a system to deliver the suggested conversational explanation and to facilitate \textit{Broad-XAI}. It consists of four components: Perception, Merkwelt Model, Actor's Model and a Behaviour Model. Arrows between components represent the direction of influence that is required by the \textit{level of explanation}, indicated by the alphanumeric character in the arrow, along with a matching colour to previous figures.}
  
  \label{fig:CognitiveProcess}
\end{figure}

The cognitive process can be regarded as the thinking, problem-solving and doing portion of the brain. In this paper, we have regarded this cognitive process to be the machine intelligence that governs an agent's behaviour and is to be explained by the social process \ref{subsection_SocialProcess}. Throughout, this paper has assumed the existence of a cognitive process based on an underlying model of AI. This is not necessarily a common model, although the components in various forms are often used. This section will briefly discuss this assumed AI model and suggest approaches towards its development. Figure \ref{fig:CognitiveProcess} provides a detailed representation of this paper's assumed AI model for the cognitive process. This model suggests four components that work together to decide on an appropriate course of action, referred to as: Perception, Merkwelt Model, Actor’s Model and a Behaviour Model. While not all of these components are required for many AI systems, they are the minimum components of a system using all levels of explanation. Included in this figure are arrows indicating the direction of influence of these components. It is these component influences that the Zeroth-to-Nth levels (as indicated) aim to explain in the last section's conversational process. While the yellow circles indicate that the components themselves, and the explanations resulting from those components, are explained by the Meta level.

Similar to agent-based approaches such as BDI-agents and Reinforcement Learning, the model assumes an external environment that the agent perceives and processes to determine the preferred behaviour. Frequently, these are regarded as a single process --- that is to say, a single algorithm may be used to perform this input to decision mapping. For instance, taking the inputs themselves as the features and processing those with a classification or regression model to determine an output or control action. When such an approach requires explanation it is typical to utilise IML, Reactive explanations, because no broader explanation is either desired or possible of such simple systems. Hence, the final output of the model is shown as being a Reactive explanation.

However, in Figure \ref{fig:CognitiveProcess} we have explicitly separated these two components into \textit{Perception} and a \textit{Behaviour Model} to emphasis that more sophisticated approaches now actively process the input features to identify higher-level concepts as the basis for the Behaviour Model's processing. For instance, the autonomous car must identify (either explicitly or implicitly) each object in its vicinity prior to any behaviour based decision is processed. By separating these components we aim to illustrate that the explanations of each of these are, while still purely reactive, aiming at providing different information. For instance, one common IML approach to explaining how a convolutional neural network identified an object, such as a road sign, is through the use of saliency maps \cite{simonyan2013deep} --- effectively pointing a finger at where it saw the object. Such an explanation is focusing on how the feature extraction layers of the network identified the object. Often this feature extraction layer is a plugable component that can be used with additional models such as decision trees \cite{li2017pedestrian} and random forests \cite{yang2015convolutional} for the subsequent mapping of behaviours. This allows additional Reactive explanations detailing how the features affected the choice of final behaviour \cite{madumal2019explainable}. This process is shown through the grey arrows through the centre of the model indicating these are all Reactive explanations.


All AI-systems have an implicit built-in set of motivations often defined by the engineer defining its success parameters or targets. This can be as simple as providing: a set of training examples; deciding on the number of clusters to be found; or, defining a reward structure. More recently many systems are incorporating an explicit attempt to model their own internal motivations. This is particularly prevalent in systems where those internal motivations have the ability to be learnt or to switch between motivations during their operation, such as emotion influenced, multi-objective, and sub-goal based approaches. Additionally, systems may explicitly provide a memory of past events that help guide future decisions, such as: utilising memories of high rewarding experiences to bias decisions \cite{ramani2019short}; gates between recurrent network iterations \cite{wang2020deep}; or, recalling past cases to guide a deliberative agent \cite{corchado2003constructing}.

In this paper, we have introduced the idea of a \textit{Merkwelt Model} to collectively refer to methods that have such built-in motivations (either implicit or explicit) influencing their behaviours. The purpose of introducing this concept is to highlight that these motivations are an important higher-level of reasoning that requires explicit explanation. In Figure \ref{fig:CognitiveProcess}, we have indicated with the red arrow that the Merkwelt Model influences the behaviour of an agent and that a Disposition explanation is required to detail the agent's motivations. For example, the autonomous car's motivation to be in the right lane based on a sub-goal to exit the road at the approaching exit. Additionally, the cognitive process illustrates that the Merkwelt Model also influences the model of the agent's future expectations of other actors in the environment. This is shown as a green arrow indicating the Nth-order explanation, as discussed in section \ref{subsection_nthorder}, illustrating the agent's motivation is actually to alter other actors' future behaviours.

The fourth component at the bottom quarter of Figure \ref{fig:CognitiveProcess} is the inclusion of an \textit{Actor's model} that is required to provide details of higher-level reasoning about external actors. Simple autonomous systems build a reactive model to the environment and this may implicitly model external actors and how they may respond to the agent's behaviours. For example, a chess playing system may assume the opposition will select its best choice. However, when operating in complex mixed domains an agent needs to explicitly model the external actor's expected behaviour based on a model of its motivations. For example, using pedestrian prediction to determine their intentions and future behaviour \cite{hirakawa2018survey, rudenko2019human, wang2019pedestrian}. These predictions directly influence the agent's choice of behaviour, and therefore, often may be required to explain its decisions. This is represented by the purple arrow between the Actor's model and the agent's Behaviour model describing the need for a Social explanation.

Similarly, the predicted behaviour of an actor may also influence the agent's own internal motivations represented by its Merkwelt Model. For instance, if the prediction is that the actor will move to be in their future path then this may cause the agent to switch focus to a safety-based avoidance objective \cite{vamplew2018human, vamplew2017steering}. Explaining changes to an agent's internal beliefs and motivations is a recognition of the cultural expectations on the agent and must be explained with Cultural explanations as indicated by the green arrow from the Actor's Model to the agent's Merkwelt Model. In reality, the changes to the Merkwelt Model from the Actor's Model and vice versa do not occur directly as indicated by Figure \ref{fig:CognitiveProcess}, but instead are changed by observing the environment and inferring. This is indicated by the inclusion of green arrows between Perception and the Merkwelt and Actor's Models.

Finally, each of the explanations generated to detail the influences of components, plus the components themselves, also need to explain how they generated their results directly. These represent Meta-explanations, which are shown as yellow circles on each component and implied within each transition arrow. As discussed in section \ref{subsection_meta}, these are IML based explanations. For example, in explaining why the autonomous car made no evasive manoeuvre due to a prediction the pedestrian would give way, a user may request Meta-explanation details of how that prediction was concluded as well as how the explanation itself was selected as the best form of justifying the actions taken.

\subsection{The Explainer's Explanation Process and Current Broad-XAI Implementations}

\begin{figure}
  \centering
  \includegraphics[trim={5.4cm 3.9cm 5.8cm 3.9cm},clip, width=\textwidth]{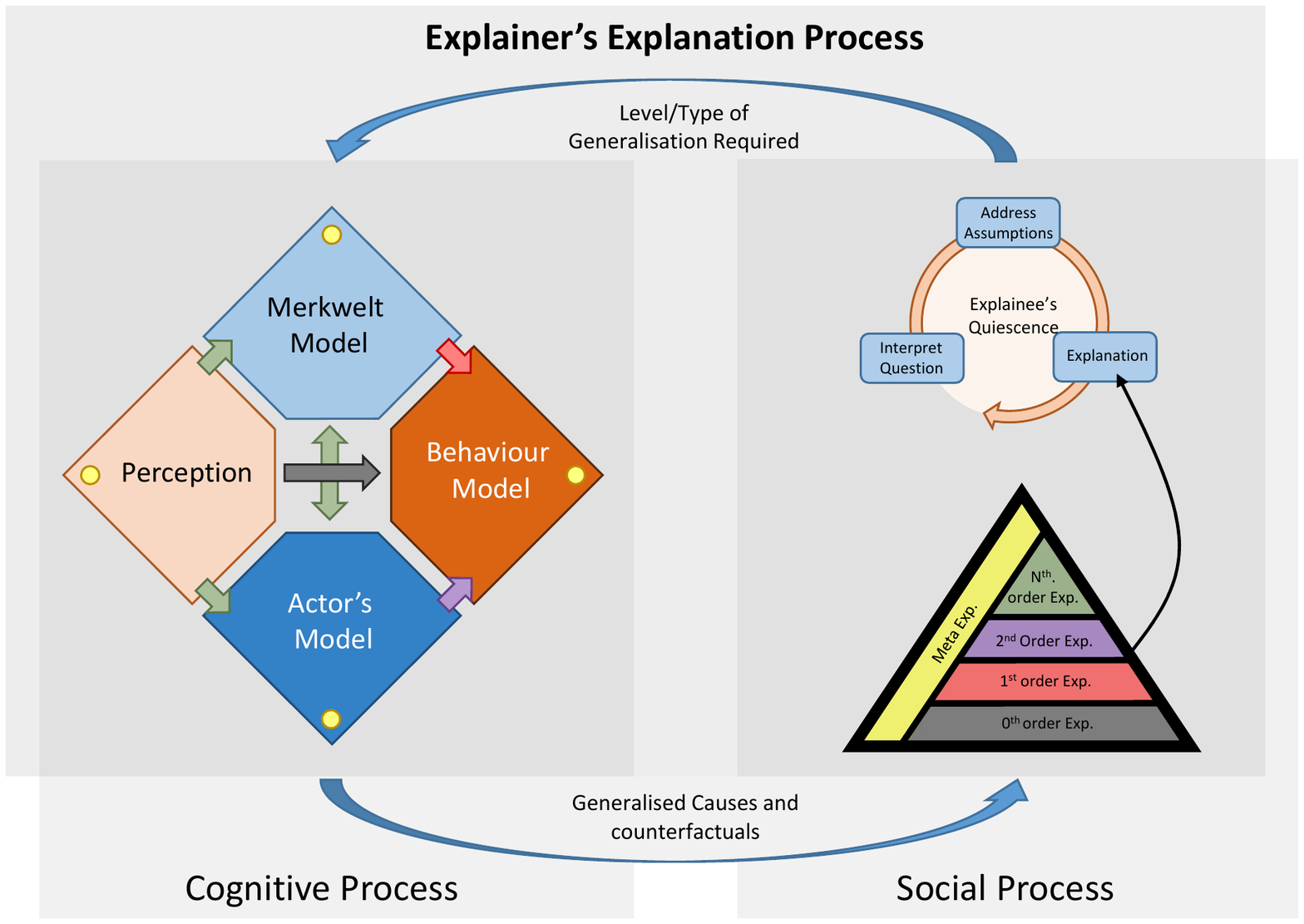} 
  \caption{\textbf{Explainer's Explanation Process}, elaborating on Figure \ref{fig:Conversation}, shows how the cognitive process links to the social process. This shows that the cognitive process determines the agent's behaviour based on the environment the agent finds itself. The social process uses the \textit{levels of explanation} to determine which of the components' influence should be used when providing an explanation.}
  \label{fig:ExplanationProcess}
\end{figure}

In the previous sub-sections, we have discussed how the agent determines the level of explanation required to resolve a user query through an iterative conversational process. This conversational process was designed as a social interface with a human's social process. Secondly, we outlined the cognitive process required of a machine intelligence to be able to support such a Broad-XAI conversational process. Figure \ref{fig:ExplanationProcess} now brings these processes together to detail the explainer's final explanation process. Here we illustrate that the agent's cognitive process, governing its behaviour, links directly to the levels of explainability introduced in this paper. When explaining the agent's behaviour the social process uses the iterative conversation process, described in sub-section \ref{subsection_SocialProcess}, which in turn maps the required level to the appropriate component and how it is influencing the agent's behaviour and its internal motivations and predictions.

For example, in explaining the cause of the accident in our motivating example a user may ask `why didn’t the car give way to the pedestrian pushing a bicycle'? In answering this question, as discussed in sub-section \ref{subsection_SocialProcess}, the system should explain at the highest relevant level, which in this case would be the agent's expectation that the pedestrian would give way. This explanation is drawing from the Actor Model's influence over the agent's behaviour, resulting in the explanation that the car `believed the person would expect the car to continue'. In the event that this explanation does not satisfy the explainee the conversational process may require an additional explanation of what the agent's motivations were when not taking avoidance measures --- such an explanation may focus on the agent's Merkwelt's choice of objective and its influence over behaviour. 

The explanation process presented here provides a structure to the generation of Broad-XAI conversational explanations combined with an underlying cognitive model of AI. The aim in presenting this process is to illustrate how several disparate explanation technologies, such as IML, Goal-driven XAI, Emotion-aware XAI, Socially-aware XAI, Utility-driven XAI etc can be brought together to explain an agent's behaviour in a holistic and user-focused manner. While this is the first philosophical drawing together of these ideas and development of levels of explainability there have already been some attempts to build such Broad-XAI approaches --- although they have not been referred to as such. In simple forms, there have been attempts to unify Reactive explanations of perception with behaviour \cite{madumal2019explainable, anderson2020mental} and attempts suggested extensions that combine these Merkwelt models \cite{cruz2019XRL, sukkerd2018toward, sukkerd2020tradeoff}.

More extensive coverage of this approach is represented using generic explanation facilities. These approaches operate entirely external to the cognitive process and build a model based on their observation of an agent's behaviour. It is based on this observed model that it provides an explanation --- hence these system tend use grey box models to predict the agent's behaviour. Two popular approaches are the Local Interpretable Model-Agnostic Explanations (LIME) \cite{ribeiro2016model} and Black Box Explanations through Transparent Approximations (BETA) \cite{lakkaraju2017interpretable}. These system do not provide an implementation anywhere near the breadth suggested as required in this paper but do often provide details of both Reactive and Disposition explanations \cite{mathews2019explainable, weitz2019you}.

One particularly notable extension to LIME attempts to implement a model that carries many similarities to the approach described in this paper. \citeauthor{neerincx2018using} \cite{neerincx2018using} describes three phases of an explanation: $\varepsilon$\textit{-generation}, $\varepsilon$\textit{-communication}, and $\varepsilon$\textit{-reception}. The $\varepsilon$\textit{-generation} phase separates the explanation of perception, behaviour and cognition creation (that we have called Merkwelt) which align with our lower two levels of explainability. The $\varepsilon$\textit{-communication} component standardises the communication with adaptive and interactive ontologies to be presented to the user, while the $\varepsilon$\textit{-reception} measures the response of the user to the explanation aligning to our utility based on quiescence. These three components align to the three components identified in our model of a conversational process in section \ref{subsection_SocialProcess}. While this work provides little philosophical or theoretical foundation to their approach and only covers the lowest two levels of explanation, it does present a clear indication that the approach suggested in this paper has potential to be implemented in real-world applications.

\section{Conclusion}

As AI is increasingly being integrated into society, research ensuring people can trust these systems has seen the emergence of eXplainable AI (XAI) as an important topic. This is particularly the case with the utilisation of black-box machine learning methods for which people have no intuition or understanding. A significant amount of XAI research, however, has been closely aligned to individual learning algorithms and have paid little attention to the need of regular people to be able to interact and understand those explanations. This paper has argued that, in order to provide acceptable and trusted explanations, a system must continually determine an explainee's contextual position through an interactive process, typically referred to as a conversation. Through this conversation, the agent progressively moves the explainee towards the point of quiescence --- a state of being quiet, where they have understood and accepted the decision. 

Furthermore, this paper has argued that AI cognition is structurally different from that of humans and that XAI systems need to go beyond cognitive interpretation alone and instead integrate technologies to achieve human-aligned conversational explanations. 
This paper's thesis has been that through the integration of technologies a Broad-XAI system can be designed to provide conversational explanations that better align the cognitive process of an AI system to that of people. 
This was accomplished by defining a set of levels for XAI explanation, Figure \ref{fig:Levels}, that cognitively aligns to the AI process. These levels were designed to specifically map to the psychological model of the human social process to address the three key areas of human explanation: contrastive explanation, attribution theory and explanation selection. There were five levels of explainability identified in this paper: Zero-order (Reactive); First-order (Disposition); Second-order (Social); N\textsuperscript{th}-order (Cultural) explanations; and, Meta (Reflective) Explanation. These levels were drawn from ethology's work in explaining animal intentionality. 

Zero-order (Reactive) explanations focus on interpreting an individual decision based on the direct information provided to it for that decision. This level is interested in the automatic reaction made based on an input set of features and does not provide an explanation for any other factors such as an agent's motivations or memory of past events. For instance, a Convolutional Neural Network (CNN) classifying types of leaves may produce a heat map or identify particular neural patterns to explain its areas of most interest when classifying a particular leaf. This level of explanation has been the focus of the majority of XAI work and is frequently referred to as Interpretable Machine learning (IML) or Transparent AI/ML.

\begin{table}[width=\linewidth,cols=8,pos=h]
\caption{A list of research fields of XAI discussed in this paper that study the explanation of AI reasoning. Each field is associated with the levels of XAI identified in the paper along with a selection of references identified.}\label{Fields of explanation}
\begin{tabular*}{\tblwidth}{p{3.5cm} p{1.0cm} p{1.4cm} p{0.8cm} p{1.0cm} p{1.2cm} p{1.0cm} p{4.0cm}}
\toprule
\multirow{2}{10em}{Fields of XAI Research} & \multicolumn{6}{l}{Levels of Explanation} & \multirow{2}{6em}{Key papers} \\ 
 & Reactive & Disposition & Social & Cultural & Reflective & Conv & \\
\midrule
Transparent AI/ML & \checkmark & & \checkmark\textsuperscript{$\dagger$} & & & & \cite{goyal2016towards, wachter2017transparent, chao2010transparent} \\
Interpretable ML (IML) & \checkmark & & \checkmark\textsuperscript{$\dagger$} & & & & \cite{abdul2018trends, adadi2018peeking, guidotti2018survey, doshi2017towards, whitby2009artificial, huysmans2011empirical, fong2017interpretable, xu2015show, zintgraf2017visualizing, SundararajanTY17} \\
Explainable Agency & & \checkmark & & & & & \cite{langley2017explainable} \\
Goal-driven XAI & & \checkmark & & & & & \cite{anjomshoaeintelligible, anjomshoae2019explainable} \\
Memory-aware XAI\textsuperscript{$\ddagger$} & & \checkmark & & & & & \cite{kroll2000grey, harbers2011explanation, harbers2011explaining, harbers2010design, cruz2019XRL, arras2017explaining, bharadhwaj2018explanations} \\
Emotion-aware XAI & & \checkmark & \checkmark & & & & \cite{kaptein2017role, rorty1978explaining, o1994explaining} \\
Socially-aware XAI\textsuperscript{$\ddagger$} & & & \checkmark & & & & \cite{mathews2019explainable, weitz2019you, hao2019emotion, sindlar2009explaining, sindlar2011programming, sindlar2008mental} \\
Culturally-aware XAI\textsuperscript{$\ddagger$} & & & & \checkmark & & & \cite{hellstrom2018understandable, wortham2017robot, dragan2013legibility, fogg2007motivating, albert2004health, kampik2019explaining} \\
Object-level Explanation & \checkmark & \checkmark & \checkmark & \checkmark &  &  & \cite{Galitsky2010ExplanationVM, galitsky2016formalizing} \\
Deceptive Explanations & & & & & \checkmark\textsuperscript{$\ast$} & &  \cite{person2019agents, van2014dynamics, van2012logic, nguyen2011asp,zlotkin1991incomplete, sakama2011logical, sakama2015formal, sakama2010many, sakama2014formal} \\
Meta-Explanation & & & & & \checkmark &  & \cite{Galitsky2010ExplanationVM, galitsky2016formalizing, pitrat2006meta,samek2019towards} \\
Utility-driven XAI\textsuperscript{$\ddagger$} & & & & & \checkmark\textsuperscript{$\ast$} & \checkmark\textsuperscript{$\ast$} & \cite{Ehsan2019OnDA, ehsan2019automated, mclaughlin1988utility} \\
Personalised Explanations & &  & & & & \checkmark\textsuperscript{$\ast$} & \cite{schneider2019personalized, tsai2019designing, quijano2017make, kirsch2017explain}\\
Interactive XAI\textsuperscript{$\ddagger$} & &  & & & & \checkmark & \cite{cerulo2011social, can2016human, elder2019living, coeckelbergh2011humans, sokol2018glass}\\
Broad-XAI\textsuperscript{$\ddagger$} & & \checkmark & \checkmark & \checkmark & \checkmark & \checkmark & \\
\bottomrule
\multicolumn{8}{l}{\textsuperscript{$\ast$} Field of research does not directly address this level but does represent a frequently included component.} \\
\multicolumn{8}{l}{\textsuperscript{$\dagger$} Not used directly in this field but can be used to partially explain the perception of other agents.} \\
\multicolumn{8}{l}{\textsuperscript{$\ddagger$} Suggested future fields of research. Work listed was published under more general fields such as XAI or IML.}
\end{tabular*}
\end{table}

First-order (Disposition) explanations focus on identifying the agent's underlying internal disposition towards the environment and other actors, and how that disposition motivated it towards a particular decision. For instance, a robot that navigates towards its primary goal location decides to take the long way around a room to avoid bumping a vase that occupied a narrow part of the more direct route. For a person to understand this behaviour the robot must explain that the decision was based on a secondary objective to avoid damaging anything while moving around. This internal motivation may also be influenced by: a memory of past events, such as the location of something it can no longer directly perceive; or, internal parameters that are set or learnable, such as emotions. Research into explanations at this level has been carried out in explainable agency, Goal-driven XAI, Memory-aware XAI, and Emotion-aware XAI. 

Second-order (Social) explanations focus on decisions based on an agent's awareness or belief of its own or other actors' mental states. This level builds on Disposition explanations and is aimed at explaining the agent's interpretation of what other actors might be thinking. In other words, if an agent's decision is affected by what it predicts others' intentions might be then it should be able to explain how its decision was changed. This level is also referred to as Social explanation as the agent is explaining a decision based on its social awareness and aligns with human reasoning in social interactions. For example, an autonomous vehicle must predict the behaviour of other road users, cars, trucks, pedestrians and animals and modify its behaviour accordingly. A Social explanation should be able to identify how the prediction of other road users' behaviour affected its decisions. The raw interpretation of predictions of other actors is frequently researched in existing Interpretable and Transparent ML, however, this only explains the prediction itself. A true Social explanation will explain how that interpretation altered its decision. Early research in this area has been discussed in both Emotion-aware XAI and Socially-aware XAI. 

N\textsuperscript{th}-order (Cultural) explanations focus on decisions made by an agent on what it has determined is expected of it culturally by other actors. This level requires an agent to not only model a prediction of other actors' behaviour but also models the expectations those actors may have about how the agent will behave. This level is also referred to as Cultural explanations as it is concerned with the cultural expectations placed on an agent's behaviour, which may differ in different locations. For example, Cultural expectations on an agent about which side of a path an agent should travel will change depending on whether it is on the Chinese mainland (right side) or Hong Kong (left side). The provision of an explanation on how such cultural factors affect an agent's decision-making has only recently started to be researched.

A concern with providing Broad-XAI systems is that if we cannot trust the original decision, how can we trust the explanation? This concern eventuates when you move away from purely IML based systems and attempt to make explanations more readily understandable and accessible to the general public. For instance, a utility-driven explanation could learn to provide understandable and readily acceptable explanations of a decision at the expense of truth. Likewise, agents may be developed that are intentionally deceptive, such as negotiation based systems. Finally, in an effort to simplify and generalise an explanation crucial details may be omitted that inadvertently mislead. To address this issue this paper also included a fifth tangential type of explanation referred to as a Meta (Reflective) explanation. A Meta-explanation is focused on explaining the process and factors that were used to generate, infer or select an explanation. 

Finally, this paper presented a model for explaining a decision or outcome through a recursive process of iterative conversation. This process starts with a generalist assumption and progressively provides greater specificity until the explainee reaches a point of acceptance, referred to as quiescence. We suggest that a mapping such as this allows researchers and developers to facilitate trust and social acceptance of AI systems as they are increasingly integrated into society through the provision of a contextually responsive system. In discussing each level we surveyed a range of research previously applied in these areas and suggested future research directions to enable progress for each area. Table \ref{Fields of explanation} presents each of the fields discussed and maps them to the levels to which they may contribute. In this way, this paper has pulled together all the disparate areas of XAI research and identified how each of these fields fit together into a broader context. In so doing, this paper aimed to promote new avenues of research into finding approaches for Broad-XAI.

\bibliographystyle{model1-num-names}

\bibliography{XAI}

\end{document}